\documentclass[12pt]{article}
\usepackage[utf8]{inputenc}

\usepackage{amsmath,amssymb,amsthm,amscd, natbib, tikz,tikz-cd}
\usepackage{wasysym, graphicx, hyperref, float}
 \usepackage[all,cmtip]{xy}
\usepackage{tikz-cd}
\usepackage{ gensymb }
 \usepackage{ textcomp, subcaption}
 \usepackage{authblk}

\usepackage{enumitem}

\newtheorem{theorem}{Theorem}[section]
\newtheorem{lemma}[theorem]{Lemma}

\newtheorem{proposition}[theorem]{Proposition}

\theoremstyle{definition}
\newtheorem{definition}[theorem]{Definition}

\theoremstyle{remark}
\newtheorem{remark}[theorem]{Remark}

\newtheorem{assumption}[theorem]{Assumption}

\newcommand{\im}{\operatorname{im}}

\newcommand{\acts}{\curvearrowright}

\mathchardef\mhyphen"2D

\newcommand{\spn}{\mathrm{span}}
\newcommand{\dis}{\displaystyle}
\newcommand{\orb}{\mathrm{Orbit}}

\newcommand{\norm}[1]{\left\lVert#1\right\rVert}

\usepackage{subcaption}

\usepackage{amsthm,amsmath,amsfonts,amssymb, mathtools}
\usepackage{xypic}

\DeclarePairedDelimiter\ceil{\lceil}{\rceil}
\DeclarePairedDelimiter\floor{\lfloor}{\rfloor}

\newcommand{\BlackBox}{\rule{1.5ex}{1.5ex}}  % end of proof
\ifdefined\proof
    \renewenvironment{proof}{\par\noindent{\bf Proof\ }}{\hfill\BlackBox\\[2mm]}
\else
    \newenvironment{proof}{\par\noindent{\bf Proof\ }}{\hfill\BlackBox\\[2mm]}
\fi

\begin{document}

\title{Reproducing Kernels and New Approaches in Compositional Data Analysis}

\author[1]{Binglin Li}
\author[2]{Jeongyoun Ahn}

\affil[1]{University of Georgia}
\affil[2]{Korea Advanced Institute of Science and Technology}

\maketitle

\begin{abstract}%   <- trailing '%' for backward compatibility of .sty file
Compositional data, such as human gut microbiomes, consist of non-negative variables whose only the relative values to other variables are available. Analyzing compositional data such as human gut microbiomes needs a careful treatment of the geometry of the data.  A common geometrical understanding of compositional data is via a regular simplex. Majority of existing approaches rely on a log-ratio or power transformations to overcome the innate simplicial geometry. In this work, based on the key observation that a compositional data are projective in nature, and on the intrinsic connection between projective and spherical geometry, we re-interpret the compositional domain as the quotient topology of a sphere modded out by a group action. This re-interpretation allows us to understand the function space on compositional domains in terms of that on spheres and to use spherical harmonics theory along with reflection group actions for constructing a \emph{compositional Reproducing Kernel Hilbert Space (RKHS)}. This construction of RKHS for compositional data will widely open research avenues for future methodology developments. In particular, well-developed kernel embedding methods can be now introduced to compositional data analysis. The polynomial nature of compositional RKHS has both theoretical and computational benefits. The wide applicability of the proposed theoretical framework is exemplified with nonparametric density estimation and kernel exponential family for compositional data. 

\end{abstract}

\section{Introduction}

Recent popularity of human gut microbiomes research has presented many data-analytic, statistical challenges \citep{calle2019statistical}. Among many features of microbiomes, or meta-genomics data, we address their \emph{compositional} nature in this work. Compositional data consist of $n$ observations of $(d+1)$ non-negative  variables whose values represent the relative proportions to other variables in the data.  Compositional data have been commonly observed in many scientific fields, such as bio-chemistry, ecology, finance, economics, to name just a few. The most notable aspect of compositional data is the restriction on their domain, specifically that the sum of the variables is fixed.  The compositional domain is not a classical vector space, but instead a (regular) simplex, that can be modeled by the simplex: 
\begin{equation}\label{eq:simplex}
\Delta^d=\left\{(x_1,\dots,x_{d+1})\in \mathbb R^{d+1} \;| \sum_{i=1}^{d+1}x_i=1,\ x_i\geq 0, \forall i \right\},
\end{equation}
which is topologically compact. The inclusion of zeros in (\ref{eq:simplex}) is crucial as most microbiomes data have a substantial number of zeros. 

Arguably the most prominent approach to handle the data on a simplex is to take a log-ratio transformation \citep{aitch86}, for which one has to consider only the open interior of $\Delta^d$, denoted by $\mathcal S^d$. Zeros are usually taken care of by adding a small number, however, it has been noted that the results of analysis can be quite dependent on how the zeros are handled \citep{lubbe2021comparison}.  \citet{micomp} pointed out ``the dangers inherent in ignoring the compositional nature of the data'' and argued that microbiome datasets must be  treated as compositions at all stages of analysis.  Recently some approaches that analyze compositional data without any transformation have been gaining popularity \citep{li2020s, rasmussen2020zero}. The approach proposed in this paper is to construct reproducing kernels of compositional data by interpreting compositional domains via projective spherical geometries. 

% Acknowledgements should go at the end, before appendices and references

\subsection{Methodological Motivation}\label{machine}

Besides the motivation from microbiomes studies, another source of inspiration for this work is the current exciting development in statistics and machine learning. In particular, the rising popularity of applying higher tensors and kernel techniques, which allows multivariate techniques to be extended to exotic structures beyond traditional vector spaces, e.g., graphs \citep{graphrkhs}, manifolds \citep{vecman} or images \citep{tensorbrain}. This work serves an attempt to construct reproducing kernel structures for compositional data, so that recent developments of (reproducing) kernel techniques from machine learning theory can be introduced to this classical field in statistics. 

The approach in this work is to model the compositional data as a group quotient of a sphere $\mathbb S^d/\Gamma$ (see (\ref{allsame})), which gives a new connection of compositional data analysis with directional statistics. The idea of representing data by using tensors and frames is not new in directional statistics \citep{ambro}, but the authors find it more convenient to construct reproducing kernels for $\mathbb S^d/\Gamma$ (whose reason is given in Section \ref{whyrkhs}).

We do want to mention that the construction of reproducing kernels for compositional data indicates a new potential paradigm for compositional data analysis: traditional approaches aim to find direct analogue of multivariate concepts, like mean, variance-covariance matrices and suitable regression analysis frameworks based on those concepts. However, finding the mean point over non-linear spaces, e.g. on manifold, is not an easy job, and in worst case scenarios, mean points might not even exist on the underlying space (e.g. the mean point of the uniform distribution on a unit circle is \emph{not} living on the circle).

In this work we take the perspective of kernel mean embedding \citep{kermean}. Roughly speaking, instead of finding the ``\emph{physical}'' point for the mean of a distribution, one could do statistics \emph{distributionally}. In other words, the mean or expectation is considered as a \emph{linear functional} on the RKHS, and this functional is represented by an actual function in the Hilbert space, which is referred to as ``kernel mean $\mathbb E[k(X,\cdot)]$''. Instead of trying to find another compositional point as the empirical mean of a compositional data set, one can construct ``kernel mean'' as a replacement of the traditional empirical mean, which is just $\sum_{i=1}^nk(X_i,\cdot)/n$. Moreover, one can also construct the analogue of variance-covariance matrix purely from kernels; in fact, \citet{fbj09} considered the gram matrix constructed out of reproducing kernels, as consistent estimators of cross-variance operators (these operators play the role of covariance and cross-variance matrices in classical Euclidean spaces).

Since we remodel compositional domain using projective/spherical geometry, compositional domain is \emph{not} treated as a vector space, but a quotient topological space $\mathbb S^d/\Gamma$. Instead of ``putting a linear structure on an Aitchison simplex''  \citep{aitch86}, or square root transformation (which is still transformed from an Aitchison simplex), we choose to ``linearize'' compositional data points by using kernel techniques (and possibly higher-tensor constructions) and one can still do ``multivariate analysis''.  Our construction in this work initiates such an attempt to introduce these recent development of kernel and tensor techniques from statistical learning theory into compositional data analysis.

\subsection{Contributions of the Present Work}
 
Our contribution in this paper is three folds. First, we propose a new geometric foundation for compositional data analysis, $\mathbb P^d_{\geq 0}$, a subspace of a full projective space $\mathbb P^d$. Based on the close connection of spheres with projective spaces, we will also describe $\mathbb P^d_{\geq 0}$ in terms of $\mathbb S^d/\Gamma$, a reflection group acting on a sphere, and the fundamental domain of this actions is the first orthant $\mathbb S^d_{\geq 0}$ (a totally different reason of using ``$\mathbb S^d_{\geq 0}$'' in the traditional approach). 

Secondly, based on the new geometric foundations of compositional domains, we propose a new nonparametric compositional density estimation by making use of the well-developed spherical density estimation theory. Furthermore, we provide a central limit theorem for integral squared errors, which leads to a goodness-of-fit test. 

Thirdly, also through this new geometric foundation, function spaces on compositional domains can be related with those on the spheres. Square integrable functions $L^2(\mathbb S^d)$ on the sphere is a focus of an ancient subject in mathematics and physics, called ``spherical harmonics''. Moreover, spherical harmonics theory also tells that each Laplacian eigenspace of $L^2(\mathbb S^d)$ is a reproducing kernel Hilbert space, and this allows us to construct reproducing kernels for compositional data points via ``orbital integrals'', which opens a door for machine learning techniques to be applied to compositional data. We also propose a compositional exponential family as a general distributional family for compositional data modeling.

\subsection{Why Projective and Spherical Geometries?}\label{whysph}

According to \cite{ai94}, ``any meaningful function of a composition must satisfy the requirement $f(ax)=f(x)$ for any $a\neq 0$.'' In geometry and topology, a space consisting of such functions is called a \emph{projective space}, denoted by $\mathbb P^d$, therefore, projective geometry should be the natural candidate to model compositional data, rather than a simplex. Since a point in compositional domains can not have opposite signs, a compositional domain is in fact a ``positive cone'' $\mathbb P_{\geq 0}^d$ inside a full projective space.

\begin{figure}
    \centering
    \includegraphics[width = .4\textwidth]{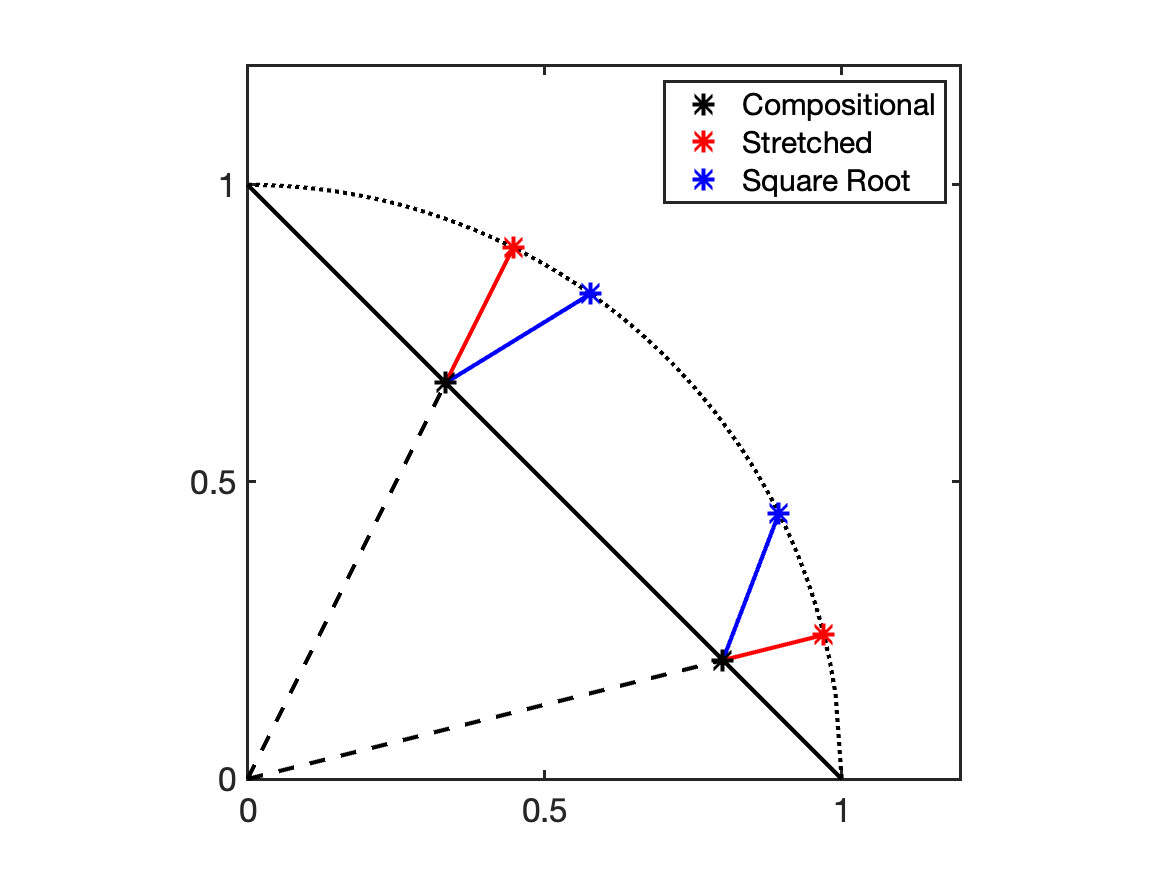}
    \caption{Illustration of the stretching action on $\Delta^1$ to $\mathbb S^1$. Note that the stretching keeps the relative compositions where the square root transformation fails to do so. }
    \label{fig:stretch}
\end{figure}

A key property of projective spaces is that stretching or shrinking the length of a vector in $\mathbb P^d$ does \emph{not} alter the point.  Thus one can stretch a point in $\Delta^d$ to a point in the first orthant sphere by dividing it by its $\ell_2$ norm. Figure \ref{fig:stretch} illustrates this stretching (``stretching'' is not a transformation from projective geometry point of view) in action. In short, projective geometry is more natural to model the compositional data according to  the original philosophy in \cite{ai94}.

However, spheres are easier to work with because mathematically speaking, the function space on spheres is a well-treated subject in spherical harmonics theory, and statistically speaking, we can connect with directional statistics in a more natural way. Our compositional domain $\mathbb P^d_{\geq 0}$ can be naturally identified with $\mathbb S^d/\Gamma$, a sphere modded out by a reflection group action. This reflection group $\Gamma$ acts on the sphere $\mathbb S^d$ by reflection, and the \emph{fundamental domain} of this action is $\mathbb S^d_{\geq 0}$ (notions of group actions, fundamental domains and reflection groups are all discussed in Section \ref{sec:sphere}). Thus our connection with the first orthant sphere $\mathbb S^d_{\geq 0}$ is a natural consequence of projective geometry and its connection with spheres with group actions, having nothing to do with square root transformations.

\subsection{Why Reproducing Kernels?}\label{whyrkhs}

As explained in Section \ref{machine}, we strive to use new ideas of tensors and kernel techniques in machine learning to propose another framework for compositional data analysis, and Section \ref{whysph} explains the new connection with spherical geometry and directional statistics. However, it is not new in directional statistics where the idea of tensors was used to represent data points \citep{ambro}. So a naive idea would be to mimic directional statistics when studying ambiguous rotations: \cite{ambro} studied how to do statistics over coset space $SO(3)/K$ where $K$ is a finite subgroup of $SO(3)$. In their case, the subgroup $K$ has to be a special class of subgroups of \emph{special} orthogonal groups, and within this class, they manage to study the corresponding tensors and frames, which gives the inner product structures of different data points. 

However, in our case, a compositional domain is $\mathbb S^d/\Gamma=O(d)\setminus O(d+1)/\Gamma$, a double coset space. Unlike \cite{ambro} that only considered $d=3$ case, our dimension $d$ is completely general; moreover, our reflection group $\Gamma$ is \emph{not} a subgroup of any special orthogonal groups, so constructions of tenors and frames in \cite{ambro} does not apply to our situation directly. 

Part of the novelty of this work is to get around this issue by making use of the reproducing kernel Hilbert Space (RKHS) structures on spheres, and ``averaging out'' the group action at the reproducing kernel level, which in return gives us a reproducing kernel structure on compositional domains. Once we have RKHS in hand, we can ``add'' and take the ``inner product'' of two data points, so our linearization strategy can also be regarded as a combination of ``the averaging approach'' and the ``embedding approach'' as in \cite{ambro}. In fact, an abstract function space together with reproducing kernels plays an increasingly important role. In below  we provide some philosophical motivations on the importance of function space over underlying data set:

\begin{itemize}
\item[(a)]Hilbert spaces of functions are naturally linear with an inner product structure. With the existence of (reproducing) kernels, data points are naturally incorporated into the function space, which leads to interesting interactions between the data set and functions defined over them. There has been a large amount of literature of embedding distributions into RKHS, e.g. \cite{disemd}, and using reproducing kernels to recover exponential families, e.g. \cite{expdual}.  RKHS has also been used to recover classical statistical tests, e.g. goodness-of-fit test in \cite{kergof}, and regression in \cite{rkrgr}. Those works do not concern the analysis of function space, but primarily focus on the data analysis on the underlying data set, but all of them are done by passing over to RKHS. This implies the increasing recognition of the importance of abstract function space with (reproducing) kernel structure.

\item[(b)] Mathematically speaking, given a geometric space $M$, the function space on $M$ can recover the underlying geometric space $M$ itself, and this principle has been playing a big role in different areas of geometry; in particular, modern algebraic geometry, following the philosophy of Grothendieck, is based upon this insight. Function spaces can be generalized to matrix valued function spaces, and this generalization gives rise to non-commutative RKHS, which is used in shape analysis in \citet{matrixvaluedker}; moreover, non-commutative RKHS is connected with free probability theory \citep{ncrkhs}, which has been used in random effects and linear mixed effects models \citep{fj19, princbulk} .

\end{itemize}

\subsection{Structure of the Paper}

We describe briefly the content of the main sections of this article:

\begin{itemize}
     \item In Section \ref{sec:sphere}, we will rebuild the geometric foundation of compositional domains by using projective geometry and spherical geometry. We will also point out that the old model using the closed simplex $\Delta^d$ is topologically the same as the new foundation. In diagrammatic way, we establish the following topological equivalence:
\begin{equation}\label{4things}
  \Delta^d\cong \mathbb P^d_{\geq 0}\cong  \mathbb S^d/\Gamma\cong\mathbb S^d_{\geq 0},  
\end{equation}

\noindent where $\mathbb S^d_{\geq 0}$ is the first orthant sphere, which is also the fundamental domain of the group action $\Gamma\acts \mathbb S^d$. All of the four spaces in (\ref{4things}) will be referred to as ``compositional domains''.

As a direct application, we propose a compositional density estimation method by using the spherical density estimation theory via a spread-out construction through the quotient map $\pi:\ \mathbb S^d\rightarrow\mathbb S^d/\Gamma$, and proved that our compositional density estimator also possesses integral square errors that satisfies central limit theorems (Theorem \ref{ourclt}), which can be used for goodness-of-fit tests.

\item Section \ref{sec:rkhs} will be devoted to constructing.  compositional reproducing kernel Hilbert spaces. Our construction relies on the reproducing kernel structures on spheres, which is given by spherical harmonics theory. \citet{wah81} constructed splines using reproducing kernel structures on $\mathbb S^2$ (2-dimensional sphere), in which she also used spherical harmonics theory in \cite{sasone}, which only treated $2$-dimensional case. Our theory deals with general $d$-dimensional case, so we need the full power of spherical harmonics theory, which will be reviewed at the beginning of Section \ref{sec:rkhs}, and then we will use spherical harmonics theory to construct compositional reproducing kernels using an ``orbital integral'' type of idea. 

\item Section \ref{sec:app} will give a couple of applications of our construction of compositional reproducing kernels. (i) The first example is the representer theorem, but with one caveat: our RKHS is finite dimensional consisting degree $2m$ homogeneous polynomials, with no transcendental functions, so linear independence for distinct data points is not directly available, however we show that when the degree $m$ is high enough, linear independence still holds. Our statement of representer theorem is not new purely from RKHS theory point of view. Our point is to demonstrate that intuitions from traditional statistical learning can still be used in compositional data analysis, with some extra care. (ii) Secondly, we construct the compositional exponential family, which can be used to model the underlying distribution of compositional data. The flexible construction will enable us to utilize the distribution family in many statistical problems such as mean tests. 

\end{itemize}

\section{New Geometric Foundations of Compositional Domains}\label{sec:sphere}

In this section, we give a new interpretation of compositional domains as a cone $\mathbb P^d_{\geq 0}$ in a projective space, based on which compositional domains can be interpreted as spherical quotients by reflection groups. This connection will yield a ``spread-out'' construction on spheres and we demonstrate an immediate application of this new approach to compositional density estimation. 

\subsection{Projective and Spherical Geometries and a Spread-out Construction}\label{sec:spread}

Compositional data consist of relative proportions of $d+1$ variables, which implies that each observation belongs to a projective space. A $d$-dimensional projective space $\mathbb P^d$ is the set of one-dimensional linear subspace of $\mathbb R^{d+1}$. A one-dimensional subspace of a vector space is just a line through the origin, and in projective geometry, all points in a line through the origin will be regarded as the same point in a projective space. Contrary to the classical linear coordinates $(x_1, \cdots,x_{d+1})$, a point in $\mathbb P^d$ can be represented by a projective coordinate $(x_1 : \cdots : x_{d+1})$, with the following property
\[
(x_1 : x_2: \cdots : x_{d+1}) = (\lambda x_1 : \lambda x_2: \cdots : \lambda x_{d+1}), ~~~\text{for any } \lambda \ne 0.
\]
It is natural that an appropriate ambient space for compositional data is \emph{non-negative projective space}, which is defined as
\begin{equation}\label{eq:proj}
\mathbb P^d_{\ge 0} = \left\{(x_1 : x_2: \cdots : x_{d+1})\in \mathbb P^d \;|  \; (x_1, x_2: \cdots : x_{d+1}) = (|x_1| : |x_2|: \cdots : |x_{d+1}|)\right \}.
\end{equation}
It is clear that the common representation of compositional data with a (closed) simplex $\Delta^d$ in (\ref{eq:simplex}) is in fact equivalent to (\ref{eq:proj}), thus we have the first equivalence:
\begin{equation}\label{projtosimp}
    \mathbb P^d_{\geq 0}\cong \Delta^d.
\end{equation}

Let $\mathbb S^d$ denote a $d$-dimensional unit sphere, defined as
\[
\mathbb S^d=\left\{(x_1,x_2,\dots, x_{d+1})\in \mathbb R^{d+1} \; | \sum_{i=1}^{d+1}x_i^2=1\right\},
\]
and let $\mathbb S^d_{\geq 0}$ denote the first orthant of $\mathbb S^d$, a subset in which all coordinates are non-negative. The following lemma states that  $\mathbb S^d_{\geq 0}$ can be a new domain for compositional data as there exists a bijective map between $\Delta^d$ and $\mathbb S^d_{\geq 0}$.

\begin{lemma} \label{compcone}
There is a canonical identification of $\Delta^d$ with $\mathbb S^d_{\geq 0}$, namely,
$$
\xymatrix{\Delta^d\ar@<.4ex>[r]^f& \mathbb S^d_{\geq 0}\ar@<.4ex>[l]^g},
$$
where $f$ is the inflation map $g$ is the contraction map, with both $f$ and $g$ being continuous and inverse to each other.
\end{lemma}

\begin{proof} 
It is straightforward to construct the inflation map $f$.  For $v \in \Delta^d$, it is easy to see that $f(v) \in \mathbb S^d_{\ge 0}$ when 
$
f(v) = v/\|v\|_2,
$
where $\|v\|_2 $ is the $\ell_2$ norm of $v$. Note that the inflation map makes sure that $f(v)$ is in the same projective space as $v$. To construct the shrinking map $g$, for $s\in \mathbb S^d_{\geq 0}$ we define
$
g(s) = s / \|s\|_1,
$ where $\|s\|_1$ is the $\ell_1$ norm of $s$ and see that $g(s)\in\Delta^d$. One can easily check that both $f$ and $g$ are continuous and inverse to each other. 

\end{proof}

Based on Lemma \ref{compcone}, we now identify $\Delta^d$ alternatively with the quotient topological space $\mathbb S^d/\Gamma$ for some group action $\Gamma$. In order to do so, we first show that the cone $\mathbb S^d_{\geq 0}$ is a strict fundamental domain of $\Gamma$, i.e., $\mathbb S^d_{\geq 0}\cong \mathbb S^d/\Gamma$. We start by defining a \emph{coordinate hyperplane} for a group. The $i$-th  coordinate hyperplane $H_i\in \mathbb R^{d+1}$ with respect to a choice of a standard basis $\{e_1,e_2,\dots, e_{d+1}\}$ is a codimension one linear subspace which is defined as
\[
H_i=\{(x_1,\dots, x_i,\dots, x_{d+1})\in \mathbb R^{d+1}:\ x_i=0\},  ~~ i = 1, \ldots, d+1.
\]
We define the reflection group $\Gamma$ with respect to coordinate hyperplanes as the follows:

\begin{definition}\label{reflect}

The reflection group $\Gamma$ is a subgroup of general linear group $GL(d+1)$ and it is generated by $\{\gamma_i, i = 1, \ldots, {d+1}\}$. Given the same basis $\{e_1,\dots, e_{d+1}\}$ for $\mathbb R^{d+1}$, the reflection $\gamma_i$ is a linear map specified via:
\[
\gamma_i:\ (x_1,\dots,x_{i-1},  x_i, x_{i+1},\dots, x_{d+1})\mapsto (x_1,\dots,x_{i-1},  -x_i, x_{i+1},\dots, x_{d+1}).
\]

\end{definition}
% Roughly speaking, our proposal is that instead of directly analyzing compositional data confined in an algebraically cumbersome space (a simplex), we map them to a unit sphere 

Note that if restricted on $\mathbb S^d$, $\gamma_i$ is an isometry map from the unit sphere  $\mathbb S^d$ to itself, which we denote by $\Gamma\acts\mathbb S^d$. Thus, one can treat the group $\Gamma$ as a discrete subgroup of the isometry group of $\mathbb S^d$. In what follows we establish that $\mathbb S^d_{\ge 0}$ is a fundamental domain of the group action $\Gamma\acts \mathbb S^d$ in the topological sense. In general, there is no uniform treatment of a fundamental domain, but we will follow the approach by \cite{bear}. To introduce a fundamental domain, let us define an \emph{orbit} first. For a point $z\in \mathbb S^d$, an orbit of the group $\Gamma$ is the following set: 
\begin{equation}\label{eq:orbit}
\orb^{\Gamma}_z=\{\gamma(z), \forall \gamma\in \Gamma\}. 
\end{equation}
Note that one can decompose $\mathbb S^d$ into a disjoint union of orbits. The size of an orbit is not necessarily the same as the size of the group $|\Gamma|$, because of the existence of a \emph{stabilizer subgroup}, which is defined as  
\begin{equation}\label{stable}
\Gamma_z=\{\gamma\in \Gamma, \gamma (z)=z\}. 
\end{equation}
The set $\Gamma_z$ forms a group itself, and we call this group $\Gamma_z$ the \emph{stabilizer subgroup} of $\Gamma$. Every element in  $\orb^{\Gamma}_z$ has isomorphic stabilizer subgroups, thus the size of $\orb^{\Gamma}_z$ is the quotient $|\Gamma|/|\Gamma_z|$, where $|\cdot|$ here is the cardinality of the sets. There are only finite possibilities for the size of a stabilizer subgroup for the action $\Gamma\acts \mathbb S^d$, and the size of stabilizer subgroups is dependent on codimensions of coordinate hyperplanes. 

\begin{definition}\label{fundomain} %\citet{bear}
Let $G$ act properly and discontinuously on a $d$-dimensional sphere, with $d>1$.  A \emph{fundamental domain} for the group action $G$ is a closed subset $F$  of the sphere such that every orbit of $G$ intersects $F$ in at least one point and if an orbit intersects with the interior of $F$ , then it only intersects $F$ at one point.
\end{definition}

A fundamental domain is \emph{strict} if every orbit of $G$ intersects $F$ at exactly one point. The following proposition identifies $\mathbb S^d_{\geq 0}$ as the quotient topological space $\mathbb S^d/\Gamma$, i.e., $\mathbb S^d_{\geq 0}=\mathbb S^d/\Gamma$.

\begin{proposition}\label{conedom}
Let $\Gamma\acts \mathbb S^d$ be the group action described in Definition \ref{reflect}, then $\mathbb S^d_{\geq 0}$ is a strict fundamental domain.
\end{proposition}

In topology, there is a natural quotient map $\mathbb S^d\rightarrow \mathbb S^d/\Gamma$. With the identification $\mathbb S^d_{\geq 0}=\mathbb S^d/\Gamma$, there should be a natural map $\mathbb S^d\rightarrow\mathbb S^d_{\geq 0}$. Now define a contraction map $c: \mathbb S^d\rightarrow \mathbb S^d_{\geq 0}$ via $(x_1,\dots,x_{d+1})\mapsto (|x_1|,\dots, |x_{d+1}|)$ by taking component-wise absolute values. Then it is straightforward to see that the $c$ is indeed the topological quotient map $\mathbb S^d\rightarrow \mathbb S^d/\Gamma$, under the identification $\mathbb S^d_{\geq 0}=\mathbb S^d/\Gamma$.

So far, via (\ref{projtosimp}), Lemma \ref{compcone} and Proposition \ref{conedom}, we have established the following equivalence:
\begin{equation}\label{allsame}
    \mathbb P^d_{\geq 0}=\Delta^d=\mathbb S^d_{\geq 0}=\mathbb S^d/\Gamma. 
\end{equation}

For the rest of the paper we will use the four characterizations of a compositional domain interchangeably.

\subsubsection{Spread-Out Construction}

Based on (\ref{allsame}), one can turn a compositional data analysis problem into one on a sphere via \emph{spread-out construction}. The key idea is to associate one compositional data point $z\in \Delta^d=\mathbb S^d_{\geq 0}$ with a $\Gamma$-orbit of data points $\orb^{\Gamma}_z\subset \mathbb S^d$ in (\ref{eq:orbit}). Formally, given a point $z\in \Delta^d$, we construct the following \emph{data set} (\emph{not necessarily a set} because of possible repetitions):
\begin{equation}\label{sprd}
    c^{-1}(z) = \left\{|\Gamma_{z'}|\ \text{copies of }z', \ \text{for}\ z'\in  \text{Orbit}_z^\Gamma \right\},  
\end{equation}

where $\Gamma_{z'}$ is the stabilizer subgroup of $\Gamma$ with respect to $z'$ in (\ref{stable}). In general, if there are $n$ observations in $\Delta^d$, the spread-out construction will create a data set with $n2^{d+1}$ observations on $\mathbb S^d$, in which observations with zero coordinates are repeated. Figure \ref{fig:kde} (a) and (b) illustrate this idea with a toy data set with $d = 2$.

\subsection{Illustration: Compositional Density Estimation}\label{compdensec}

\begin{figure}[ht]
\centering
\begin{subfigure}[b]{0.45\textwidth}
\includegraphics[width = \textwidth]{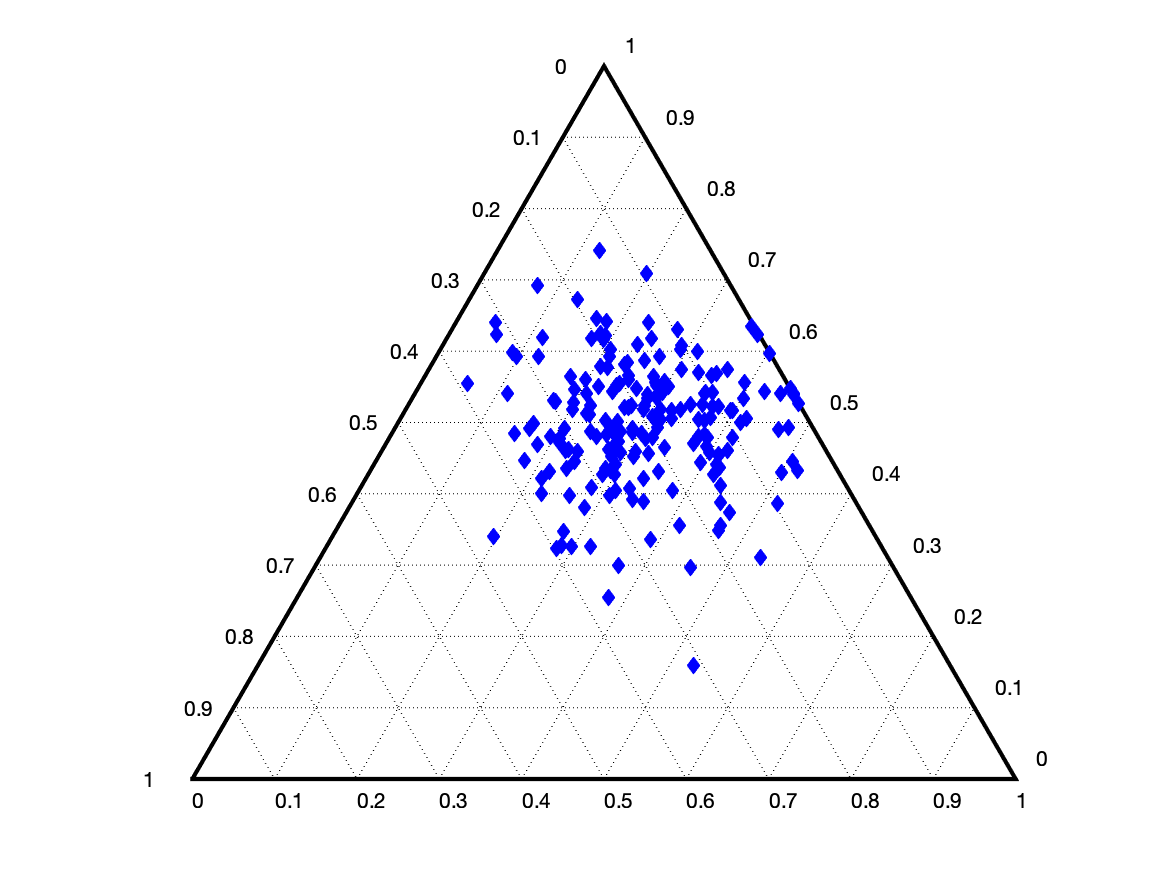}
\caption{Compositional data on $\Delta^2$}\
\end{subfigure}
\begin{subfigure}[b]{0.49\textwidth}
\includegraphics[width = \textwidth]{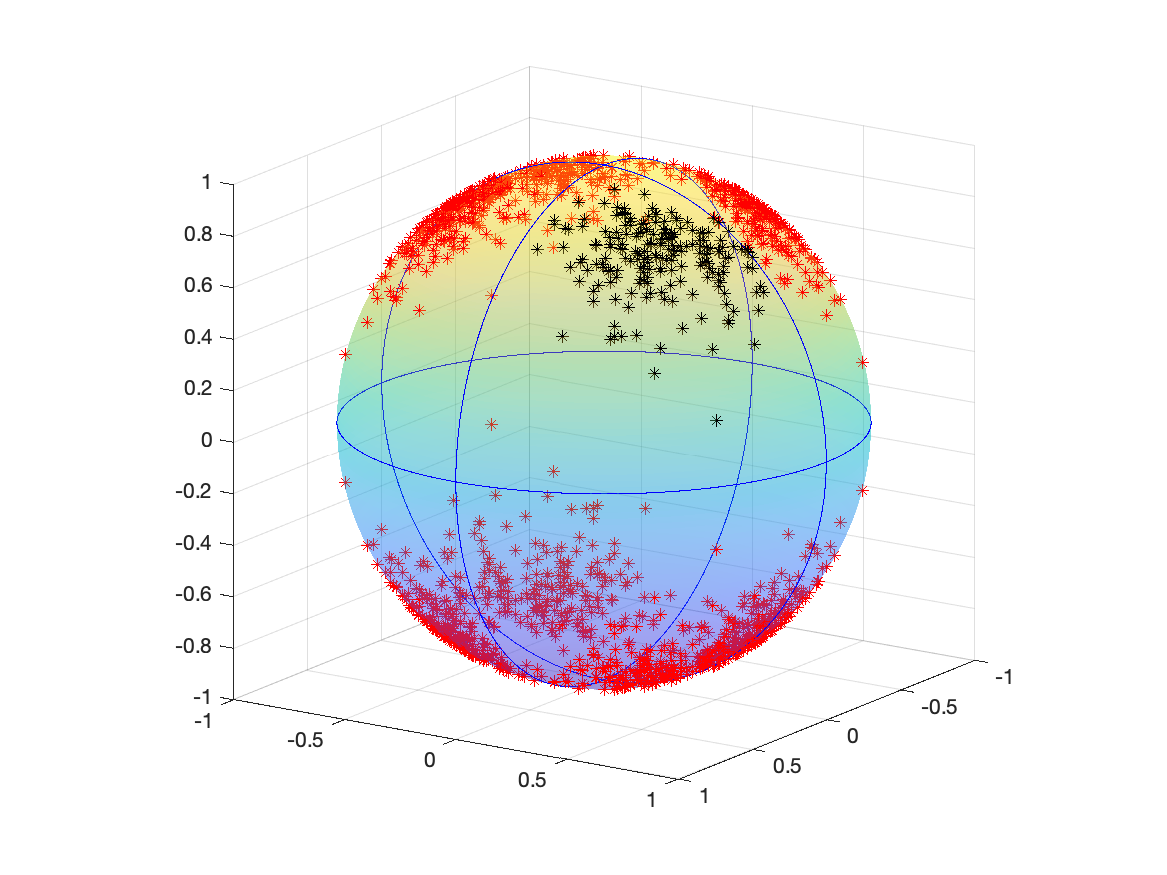}
\caption{``Spread-out'' data on $\mathbb S^2$}
\end{subfigure}\\
\begin{subfigure}[b]{0.50\textwidth}
\includegraphics[width = \textwidth]{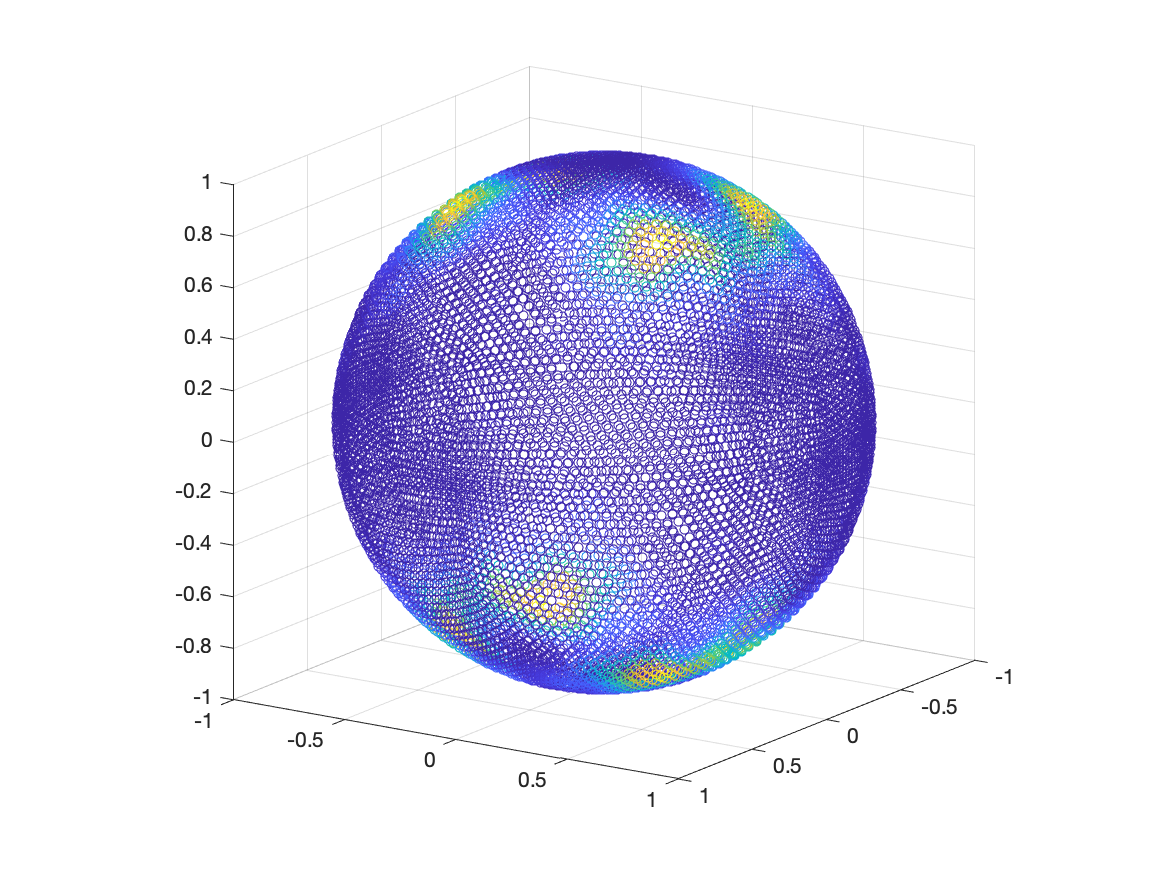}
\caption{Density estimate on $\mathbb S^2$}
\end{subfigure}
\begin{subfigure}[b]{0.40\textwidth}
\includegraphics[width = \textwidth]{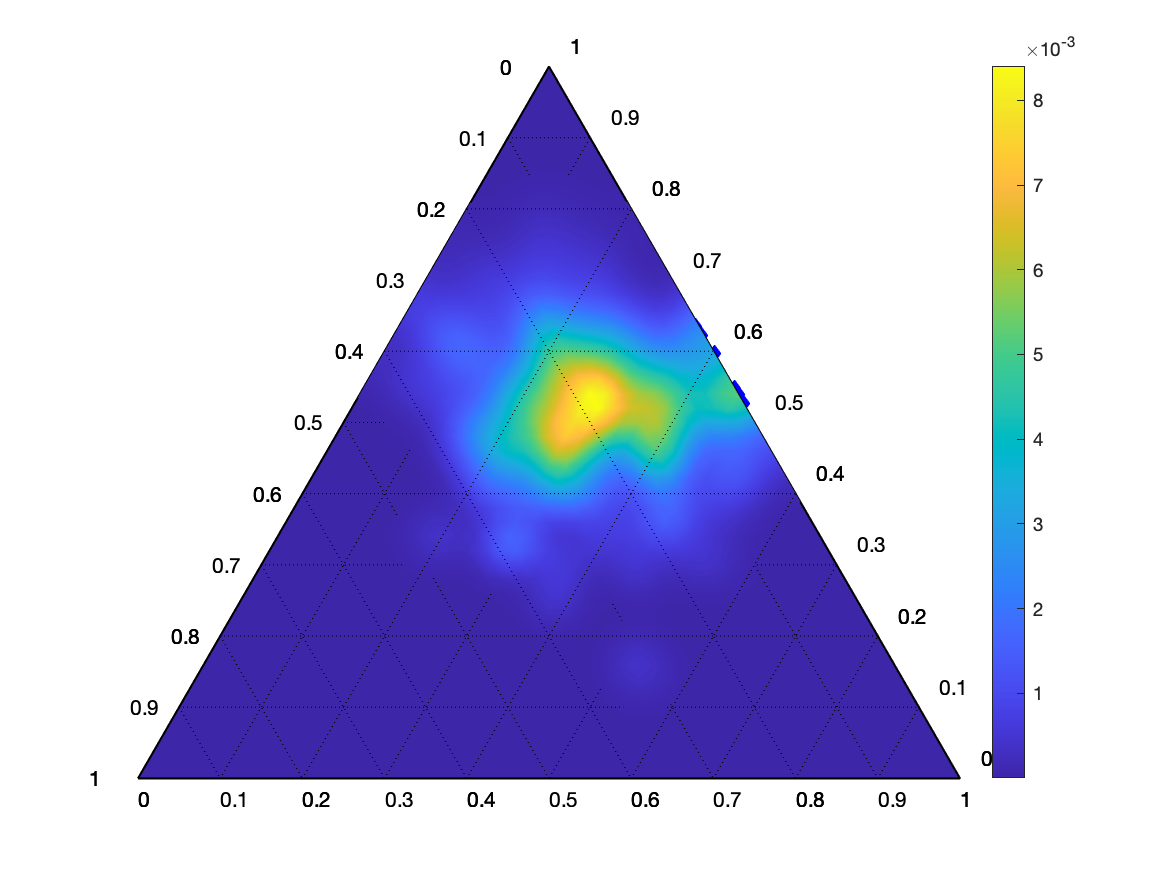}
\caption{``Pulled-back'' estimate on $\Delta^2$}
\end{subfigure}
\caption{Toy compositional data on the simplex $\Delta^2$ in (a) are spread out to a sphere $\mathbb S^2$ in (b). The density estimate on $\mathbb S^2$ in (c) are pulled back to $\Delta^2$ in (d).}\label{fig:kde}
\end{figure}

The spread-out construction in (\ref{sprd}) provides a new intimate relation between directional statistics and compositional data analysis. Indeed, this construction produces a directional data set out of a compositional data set, then we can literally transform a compositional data problem into a directional statistics problem via this spread-out construction. For example, we can perform compositional independence/uniform tests by doing directional independence/uniform tests \citep{sobind, sobuni} through spread-out constructions.

In this section, we will give a new compositional density estimation framework by using spread-out constructions. In directional statistics, density estimation for spherical data has a long history dating back to the late 70s in \cite{expdirect}. In the 80s, \cite{hallden} and \cite{baiden} established systematic framework for spherical density estimation theory. Spherical density estimation theory became popular later partly because its integral squared error (ISE) is close dly related with goodness of fit test as in \cite{chineseclt} and a recent work \cite{portclt}. 

The rich development in spherical density estimation theory will yield a compositional density framework via spread-out constructions.  In the following we apply this idea to nonparametric density estimation for compositional data. Instead of directly estimating the density on $\Delta^d$, one can perform the estimation with the spread-out data on $\mathbb S^d$, from which a density estimate for compositional data can be obtained. 

Let $p(\cdot)$ denote a probability density function of a random vector $Z$ on $\mathbb S^d_{\geq 0}$, or equivalently on $\Delta^d$. The following proposition gives a form of the density of the spread-out random vector $\Gamma(Z)$ on the whole sphere $\mathbb S^d$.

\begin{proposition}\label{induced}
Let $Z$ be a random variable on $\mathbb S^d_{\geq 0}$ with probability density $p(\cdot)$, then the induced random variable $\Gamma(Z)=\{\gamma(Z)\}_{\gamma\in \Gamma}$, has the following density  $\tilde{p}(\cdot)$ on $\mathbb S^d$:
\begin{equation}\label{cshriek}
\tilde{p}(z)=\frac{|\Gamma_z|}{|\Gamma|} p(c(z)), \ z\in \mathbb S^d,
\end{equation}
where $|\Gamma_z|$ is the cardinality of the stabilizer subgroup $\Gamma_z$ of $z$. 
\end{proposition}

Let $c_*$ denote the analogous operation for functions to the contraction map $c$ that applies to data points. It is clear that given a probability density $\tilde p$ on $\mathbb S^d$, we can obtain the original density on the compositional domain via the ``pull back'' operation $c_*$: 
\[
p(z) = c_*(\tilde p)(z)=\sum_{x\in c^{-1}(z)}\tilde p(x),  ~~  z\in \mathbb S^d_{\geq 0}. 
\]  

Now consider estimating density on $\mathbb S^d$ with the spread-out data. Density estimation for data on a sphere has been well studied in directional statistics \citep{hallden, baiden}. For $x_1, \ldots x_n \in \mathbb S^d$, a density estimate for the underlying density is 
\[
\hat{f}_n(z)=\frac{c_h}{n}\sum_{i=1}^nK\left(\frac{1-z^T x_i}{h_n}\right),\ z\in \mathbb S^d,
\]
where $K$ is a kernel function that satisfies common assumptions in Assumption \ref{kband}, and $c_h$ is a normalizing constant. Applying this to the spread-out data $c^{-1}(x_i)$, $i = 1, \ldots, n$, we have a density estimate of $\tilde p (\cdot) $ defined on $\mathbb S^d$:
\begin{equation}\label{fhattilde}
\hat{f}^{\Gamma}_n(z)=\dis \frac{c_h}{n|\Gamma|}\sum_{1\leq i\leq n,\gamma\in \Gamma}K\left(\frac{1-z^T \gamma( x_i)}{h_n}\right), \ z\in \mathbb S^d, 
\end{equation}
from which a density estimate on the compositional domain is obtained by applying $c_*$.  That is, 
\[
\hat{p_n}(z)=c_*\hat{f}^{\Gamma}_n(z)=\sum_{x\in c^{-1}(z)}\hat{f}^\Gamma_n(x),\ \ z\in \mathbb S^d_{\geq 0}. 
\]
Figure \ref{fig:kde} (c) and (d) illustrate this density estimation process with a toy example. 

The consistency of the spherical density estimate $\hat f_n$ is established by \cite{ chineseclt, portclt}, where it is shown that the integral squared error (ISE) of $\hat f_n$, $\int_{\mathbb S^d} (\hat f_n - f)^2 dz$ follows a central limit theorem. It is straightforward to show that the ISE of the proposed compositional density estimator $\hat{p}_n$ on the compositional domain also asymptotically normally distributed by CLT. However, the CLT of ISE for spherical densities in \cite{chineseclt} contains an unnecessary finite support assumption on the density kernel function $K$ (very different from reproducing kernels); although in \cite{portclt} such finite support condition is dropped, their result was on directional-linear data, and their proof does not directly applies to the pure directional context. For the readers' convenient, we will provide the proof for the CLT of ISE for both compositional and spherical data, without the finite support condition as in \cite{chineseclt}

\begin{theorem}\label{ourclt}
CLT for ISE holds for both directional and compositional data under the mild conditions (H1, H2 and H3) in Section \ref{cltprf}, without the finite support condition on density kernel functions $K$.
\end{theorem}

 The detail of the proof of THeorem \ref{ourclt} plus the statements of the technical conditions can be found Section \ref{cltprf}.

\section{Reproducing Kernels of Compositional Data}\label{sec:rkhs}

We will be devoted to construct reproducing kernel structures on compositional domains, based on the topological re-interpretation of $\Delta^d$ in Section \ref{sec:sphere}. The key idea is that based on the quotient map $\pi:\ \mathbb S^d\rightarrow \mathbb S^d/\Gamma=\Delta^d$, we can use function spaces on spheres to understand function spaces on compositional domains. Moreover, we can construct reproducing kernel structures of a compositional domain $\Delta^d$ based on those on $\mathbb S^d$.

The reproducing kernel was first introduced in 1907 by Zaremba when he studied boundary value problems for harmonic and biharmonic functions, but the systematic development of the subject was finally done in the early 1950s by \citet{aron50}. Reproducing kernels on $\mathbb S^d$ were essentially discovered by Laplace and Legendre in the 19th centuary, although the reproducing kernels on spheres were called \emph{zonal spherical functions} at that time. Both spherical harmonics theory and RKHS have found applications in theoretical subjects like functional analysis, representation theory of Lie groups and quantum mechanics. In statistics, the successful application of RKHS in spline models by \citet{wah81} popularized RKHS theory for $\mathbb S^d$. In particular, they used spherical harmonics theory to construct an RKHS on $\mathbb S^2$. Generally speaking, for a fixed topological space $X$, there exists (and one can construct) multiple reproducing kernel Hilbert spaces on $X$; In their work, an RKHS on $\mathbb S^2$ was constructed by considering a \emph{subspace} of $L^2(\mathbb S^2)$ under a finiteness condition, and the reproducing kernels were also built out of zonal spherical functions. Their work is motivated by studying spline models on the sphere, while our motivation has nothing to do with spline models of any kind. In this work we consider reproducing structures on spheres which are \emph{different} from the one in \cite{wah81}, but we share the same building blocks, which is spherical harmonics theory.

Evolved from the re-interpretation of a compositional domain $\Delta^d$ as $\mathbb S^d/\Gamma$, we will construct reproducing kernels of compositional by using reproducing kernel structures on spheres. Since spherical harmonics theory gives reproducing kernel structures on $\mathbb S^d$, and a compositional domain $\Delta^d$ are topologically covered by spheres with their deck transformations group $\Gamma$. Thus naturally we wonder (i) whether function spaces on $\Delta^d$ can identified with the subspaces of $\Gamma$-invariant functions on $\mathbb S^d$, and (ii) whether one might ``build'' $\Gamma$-invariant kernels out of spherical reproducing kernels, and hope that the $\Gamma$-invariant kernels can play the role of ``reproducing kernels''  on $\Delta^d$. It turns out that the answers for both (i) and (ii) are positive (see Remark \ref{dreami} and Theorem \ref{reprcomp}). The discovery of reproducing kernel structures on $\Delta^d$ is crucially based on the reinterpretation of compositional domains via projective and spherical geometries in Section \ref{sec:sphere}.

By considering $\Gamma$-invariant objects in spherical function spaces we managed to construct reproducing kernel structures for compositional domains, and  compositional reproducing Hilbert spaces. Although compositional RKHS was first considered as a candidate ``inner product space'' for data points to be mapped into, the benefit of working with RKHS goes far beyond than this, due to exciting development of kernel techniques in machine learning theory that can be applied to compositional data analysis as is mentioned in Section \ref{machine}. This gives a new chance to construct a new framework for compositional data analysis, in which we ``upgrade'' compositional data points as functions (via reproducing kernels), and the classical statistical notions, like means and variance-covariances, will be ``upgraded'' to linear functionals and linear operators over the functions space. Traditionally important statistical topics such as dimension reduction, regression analysis, and many inference problems can be re-addressed in the light of this new ``kernel techniques''.

\subsection{Recollection of Basic Facts from Spherical Harmonics Theory}

We give a brief review of the theory of spherical harmonics in the following. See \citet{atkinson2012spherical} for a general introduction to the topic.  In classical linear algebra, a finite dimensional linear space with a linear map to itself can be decomposed into direct sums of eigenspaces. Such a phenomenon still holds for $L^2(\mathbb S^d)$ with Laplacians being the linear operator to itself. Recall that the Laplacian operator on a function $f$ with $d+1$ variables is 
\[
\dis\Delta f=\sum_{i=1}^{d+1}\frac{\partial ^2 f}{\partial x_i^2}.
\]
Let $\mathcal H_i$ be the $i$-th eigenspace of the Laplacian operator. It is known that $L^2(\mathbb S^d)$ can be orthogonally decomposed as
\begin{equation}\label{L2}
  L^2(\mathbb S^d)=\bigoplus_{i=1}^{\infty}\mathcal H_i,
\end{equation}
where the orthogonality is endowed with respect to the inner product in $L^2(\mathbb S^d)$: $\langle f, g \rangle = \dis\int_{\mathbb S^d} f \bar{g}$. 

Let $\mathcal P_{i}(d+1)$ be the space of homogeneous polynomials of degree $i$ in $d+1$ coordinates on $\mathbb S^d$. A homogeneous polynomial is a polynomial whose terms are all monomials of the same degree, e.g., $\mathcal P_4(3)$ includes $xy^3 + x^2yz$. Further, let $ H_{i}(d+1)$ be the space of homogeneous harmonic polynomials of degree $i$ on $\mathbb S^d$, i.e.,
\begin{equation}\label{eq:harmonic}
H_{i}(d+1)=\{P\in \mathcal P_{i}(d+1)|\; \Delta P=0\}.
\end{equation}
For example,  $x^3y + xy^3 - 3xyz^2$ and $x^2 - 6x^2 y^2 + y^4$ are members of $H_4(3)$. 

Importantly, the spherical harmonic theory has established that each eigenspace $\mathcal H_{i}$ in \eqref{L2} is indeed the same space as $ H_i(d+1)$. This implies that any function in $L^2(\mathbb S^d)$ can be approximated by an accumulated direct sum of orthogonal homogeneous harmonic polynomials. The following well-known proposition further reveals that the Laplacian constraint in (\ref{eq:harmonic}) is not necessary to characterize the function space on the sphere. 

\begin{proposition}\label{accudirct}
Let $\mathcal P_{m}(d+1)$ be the space of degree $m$ homogeneous polynomial on $d+1$ variables on the unit sphere and $\mathcal H_i$ be the $i$th eigenspace of $L^2(\mathbb S^d)$. Then 
\[
\mathcal P_{m}(d+1)=\dis\bigoplus_{i=\ceil{m/2}-\floor{m/2}}^{\floor{m/2}}\mathcal H_{2i},
\]
where $\ceil{\cdot}$ and $\floor{\cdot}$ stand for round-up and round-down integers respectively. 
\end{proposition}

From Proposition \ref{accudirct}, one can see that any $L^2$ function on $\mathbb S^d$ can be approximated by homogeneous polynomials. An important feature of spherical harmonics theory is that it gives reproducing structures on spheres, and now we will recall this fact. For the following discussion, we will fix a Laplacian eigenspace $\mathcal H_i$ inside $L^2(\mathbb S^d)$, so $\mathcal H_i$ is a finite dimensional Hilbert space on $\mathbb S^d$; such a restriction on a single piece $\mathcal H_i$ is necessary because the entire Hilbert space $L^2(\mathbb S^d)$ does not have a reproducing kernel given that the Delta functional on $L^2(\mathbb S^d)$ is \emph{not} a bounded functional\footnote{At first sight, this might seem to contradict the discussion on splines on $2$-dimensional spheres in \cite{wah81}, but a careful reader can find that a finiteness constraint was imposed there, and it was \emph{never} claimed that $L^2(\mathbb S^2)$ is a RKHS. That is, their RKHS on $\mathbb S^2$ is a subspace of $L^2(\mathbb S^2)$.}. 

\subsection{Zonal Spherical Functions as Reproducing Kernels in $\mathcal H_i$}
On each Laplacian eigenspace $\mathcal H_i$ inside $L^2(\mathbb S^d)$ on general $d$-dimensional spheres, we define a linear functional $L_x$ on $\mathcal H_i$, such that for each $Y\in \mathcal H_i$, $L_x(Y)=Y(x)$ for a fixed point $x\in \mathbb S^d$. General spherical harmonics theory tells us that there exists $k_i(x,t)$ such that:
$$
L_x(Y)=Y(x)=\dis\int_{\mathbb S^d} Y(t)k_i(x,t)dt,\ x\in \mathbb S^d;
$$

\noindent this function $k_i(x,t)$ is the representing function of the functional $L_x(Y)$, and classical spherical harmonics theory refers to the function $k_i(x,t)$ as the \emph{zonal spherical function}, and furthermore, they are actually ``reproducing kernels'' inside $\mathcal H_i\subset L^2(\mathbb S^d)$ in the sense of \cite{aron50}.    Another way to appreciate spherical harmonics theory is that it tells that each Laplacian eigenspace $\mathcal H_i\subset L^2(\mathbb S^d)$ is actually a reproducing kernel Hilbert space on $\mathbb S^d$, a special case when $d = 2$  was used \cite{wah81}. 

Let us recollect some basic facts of zonal spherical functions for readers' convenience in the next Proposition.  One can find their proofs in almost any modern spherical harmonics references, in particular in \citet[Chapter IV]{stein71}:
\begin{proposition}\label{reprsph}
The following properties hold for the zonal spherical function $k_i(x,t)$, which is also the reproducing kernel inside $\mathcal H_i\subset L^2(\mathbb S^d)$ with dimension $a_i$. 
\begin{itemize}
\item[(a)]For a choice of orthonormal  basis $\{Y_1, \dots, Y_{a_i}\}$ in $\mathcal H_i$, we can express the kernel $k_i(x,t)=\dis\sum_{i=1}^{a_i}\overline{Y_{i}(x)}Y_{i}(t)$, but $k_i(x,t)$ does not depend on choices of basis.

\item[(b)]$k_i(x,t)$ is a real-valued function and symmetric, i.e., $k_i(x,t)=k_i(t,x)$.

\item[(c)]For any orthogonal matrix $R\in O(d+1)$, we have $k_i(x,t)=k_i(Rx, Rt)$.

\item[(d)] $k_i(x,x)=\dis\frac{a_i}{\mathrm{vol}(\mathbb S^d)}$ for any point $x\in \mathbb S^d$. 

\item[(e)]$k_i(x,t)\leq \dis\frac{a_i}{\mathrm{vol}(\mathbb S^d)}$ for any $x,\ t\in \mathbb S^d$.
\end{itemize}

\end{proposition}

\begin{remark}
\normalfont The above proposition ``\emph{seems}'' obvious from traditional perspectives, as if it could be found in any textbook, so readers with rich experience with RKHS theory might think that we are stating something trivial. However, we want to point out two facts. 

\begin{itemize}
    \item [(1)] Function spaces over underlying spaces with different topological structures behave very differently. Spheres are compact with no boundary, and their function spaces have Laplacian operators whose eigenspaces and finite dimensional, which possesses reproducing kernels structures inside finite dimensional eigenspaces. These coincidences are not expected to happen over other general topological spaces.
    
    \item[(2)]Relative to classical topological spaces whose RKHS were used more often, e.g. unit intervals or vector spaces, spheres are more ``exotic'' topological structures (simply connected space, but with nontrivial higher homotopy groups), while intervals or vector spaces are contractible with trivial homotopy groups. One way to appreciate spherical harmonics theory is that classical ``naive'' expectations can still happen on spheres.
\end{itemize}

\end{remark}

In the next subsection we discuss the corresponding function space in the compositional domain $\Delta^d$. 

\subsection{Function Spaces on Compositional Domains}\label{sec:fcomp}

With the identification $\Delta^d=\mathbb S^d/\Gamma$,  the functions space $L^2(\Delta^d)$ can be identified with $L^2(\mathbb S^d/\Gamma)$, i.e.,  $L^2(\Delta^d)=L^2(\mathbb S^d/\Gamma)$. The function space $L^2(\mathbb S^d)$ is well understood by spherical harmonics theory as above, so we want to relate $L^2(\mathbb S^d/\Gamma)$ with  $L^2(\mathbb S^d)$ as follows. Notice that a function $h\in L^2(\mathbb S^d/\Gamma)$ is a map from $\mathbb S^d/\Gamma$ to (real or complex) numbers. Thus a natural associated function $\pi^*(h)\in L^2(\mathbb S^d)$ is given by the following composition of maps:
$$
\pi\circ h:\ \ \xymatrix{\mathbb S^d\ar[r]^-{\pi}&\mathbb S^d/\Gamma\ar[r]^{\ \ h}& \mathbb C}.
$$
Therefore, the composition $\pi\circ h=\pi^*(h)\in L^2(\mathbb S^d)$ gives rise to a natural embedding of the function space of compositional domains to that of a sphere $\pi^*:\ L^2(\mathbb S^d/\Gamma)\rightarrow L^2(\mathbb S^d)$.

The embedding $\pi^*$ identifies the Hilbert space of compositional domains as a subspace of the Hilbert space of spheres. A natural question is how to characterize the subspace in $L^2(\mathbb S^d)$ that corresponds to functions on compositional domains.  The following proposition states that  $f\in \im(\pi^*)$ if and only if $f$ is constant on fibers of the projection map $\pi:\ \mathbb S^d\rightarrow \mathbb S^d/\Gamma$, almost everywhere.  In other words, $f$ takes the same values on all $\Gamma$ orbits, i.e., on the set of points which are connected to each other by ``sign flippings''.

\begin{proposition}\label{compfun}
The image of the embedding $\pi^*: L^2(\mathbb S^d/\Gamma)\rightarrow L^2(\mathbb S^d)$ consists of functions $f\in L^2(\mathbb S^d)$ such that up to a measure zero set, is constant on $\pi^{-1}(x)$ for every $x\in \mathbb S^d/\Gamma$, where $\pi$ is the natural projection  $\mathbb S^d\rightarrow \mathbb S^d/\Gamma$.
\end{proposition}

We call a function $f\in L^2(\mathbb S^d)$ that lies in the image of the embedding $\pi^*$ a \emph{$\Gamma$-invariant function}. Now we construct the contraction map $\pi_{*}: L^2(S^d)\rightarrow L^2(S^d/\Gamma)$ and this map will descend every function on spheres to a function on compositional domains.  To construct $\pi_*$, it suffices to associate a $\Gamma$-invariant function to every function in $L^2(\mathbb S^d)$. For a point $z\in \mathbb S^d$ and a reflection $\gamma\in \Gamma$, a point $\gamma(z)$ lies in the set $\orb_z^\Gamma$ which is defined in (\ref{eq:orbit}). Starting with a function $f\in L^2(\mathbb S^d)$, we will define the associated $\Gamma$-invariant function $f^{\Gamma}$ as follows:

\begin{proposition}
Let  $f$ be a function in $L^2(\mathbb S^d)$. Then the following $f^\Gamma$ 
\begin{equation}\label{eq:invfun}
f^{\Gamma}(z)=\dis\frac{1}{|\Gamma|}\sum_{\gamma\in \Gamma}f(\gamma(z)), ~~~ z \in \mathbb S^d,
\end{equation}
is a $\Gamma$-invariant function.
\end{proposition}
\begin{proof}
Each fiber of the projection map $\pi:\ \mathbb S^d\rightarrow \mathbb S^d/\Gamma$ is $\orb_z^\Gamma$ for some $z$ in the fiber.  For any other point $y$ on the same fiber with $z$ for the projection $\pi$, there exists a reflection $\gamma\in \Gamma$ such that $y=\gamma (z)$. Then this proposition follows from the identity $f^{\Gamma}(z)=f^{\Gamma}(\gamma(z))$, which can be easily checked.

\end{proof}

The contraction map $ f\mapsto f^{\Gamma}$ on spheres naturally gives the following map 
\begin{equation}\label{lowerstar}
  \pi_*:\ L^2(\mathbb S^d)\rightarrow L^2(\mathbb S^d/\Gamma),\  \text{with}\ f\mapsto f^{\Gamma}
\end{equation}

\begin{remark}
\normalfont Some readers might argue that each element in an $L^2$ space is a \emph{function class} rather than a function, so in that sense $\pi_*(f)=f^{\Gamma}$ is not well-defined, but note that each element in $L^2(\mathbb S^d)$ can be approximated by polynomials, and the $\pi_*$ which is well defined on individual polynomial, will induce a well defined map on function classes. 
\end{remark}

\begin{theorem}\label{invfunsp}
This contraction map $\pi_*:\ L^2(\mathbb S^d)\rightarrow L^2(\mathbb S^d/\Gamma)$, as defined in (\ref{lowerstar}),  has a section given by $\pi^*$, namely the
 composition $\pi_*\circ \pi^*$ induces the identity map from $L^2(\mathbb S^d/\Gamma)$ to itself.  In particular, the contraction map $\pi_*$ is a surjection.
\end{theorem}

\begin{proof}
One way to look at the relation of the two maps $\pi_*$ and $\pi_*$ is through the diagram $\xymatrix{L^2(\mathbb S^d/\Gamma)\ar@<.35ex>[r]^-{\pi^*}&L^2(\mathbb S^d)\ar@<.35ex>[l]^-{\pi_*}}$.  The image of $\pi^*$ consists of $\Gamma$-invariant functions in $L^2(\mathbb S^d)$. 

Conversely, given a $\Gamma$-invariant function $g\in L^2(\mathbb S^d)$, the map $g\mapsto g^{\Gamma}$ is an identity map, i.e., $g=g^{\Gamma}$, thus the theorem follows.

\end{proof}

\begin{remark}\label{dreami}
\normalfont Theorem \ref{invfunsp} identifies functions on compositional domains as $\Gamma$-invariant functions in $L^2(\mathbb S^d)$. For any function $f\in L^2(\mathbb S^d)$, we can produce the corresponding $\Gamma$-invariant function $f^{\Gamma}$ by (\ref{eq:invfun}).  More importantly, we can ``recover'' $L^2(\Delta^d)$ from $L^2(\mathbb S^d)$, without losing any information. This  allows us to construct reproducing kernels of $\Delta^d$ from $L^2(\mathbb S^d)$ in  Section \ref{sec:rkhsc}.
\end{remark}

%Corollary \ref{idem} implies that the contraction map $\pi_*$ is essentially a \emph{projection}, which is an incarnation of ``projection matrix'' in classical linear regression theory.

\subsection{ Further Reduction to Homogeneous Polynomials of Even Degrees}\label{redsurg}

In this section we provide a further simplification of the homogeneous polynomials in the finite direct sum space $\bigoplus_{i=0}^{m}\mathcal H_i$.  Proposition \ref{accudirct} tells us that if $m$ is even, then $\mathcal P_{m}(d+1)=\bigoplus_{i=0}^{m/2}\mathcal H_{2i}$, and that if $m$ is odd then $\mathcal P_{m}(d+1)=\bigoplus_{i=0}^{(m-1)/2}\mathcal H_{2i+1}$, where $\mathcal P_{m}(d+1)$ is the space of degree $m$ homogeneous polynomials in $d+1$ variables. In either of the cases ($m$ being even or odd), the degree of the homogeneous polynomials $m$ is the same as the $\max\{2i,\ \ceil{m/2}-\floor{m/2}\leq i\leq \floor{m/2}\}$. Therefore we can decompose the finite direct sum space $\bigoplus_{i=0}^{m}\mathcal H_i$ into the direct sum of two homogeneous polynomial spaces:
$$
\dis\bigoplus_{i=0}^{m}\mathcal H_i=\mathcal P_{m}(d+1)\bigoplus \mathcal P_{m-1}(d+1).
$$
However we will show that any monomial of odd degree term will collapse to zero by taking its $\Gamma$-invariant, thus only one piece of the above homogeneous polynomial space will ``survive'' under the contraction map $\pi_*$.  This will further simplify the function space, which in turn facilitates an easy computation. 

Specifically, when working with accumulated direct sums $\bigoplus_{i=0}^{m}\mathcal H_i$ on spheres, not every function is a meaningful function on $\Delta^d=\mathbb S^d/\Gamma$, e.g., we can find a nonzero function $f\in \bigoplus_{i=0}^{m}\mathcal H_i$, but $f^{\Gamma}=0$.  In fact, all of the odd pieces of the eigenspace $\mathcal H_m$ with $m$ being odd do not contribute to anything to $L^2(\Delta^d)=L^2(\mathbb S^d/\Gamma)$.  In other words, the accumulated direct sum $\bigoplus_{i=0}^m\mathcal H_{2i+1}$ is ``killed'' to zero under  $\pi_*$, as shown by the following Lemma:

\begin{lemma}\label{killodd}
For every monomial $\prod_{i=1}^{d+1}x_i^{\alpha_i}$ (each $\alpha_i\geq 0$),  if there exits $ k$ with $\alpha_k$ being odd, then the monomial $\prod_{i=1}^{d+1}x_i^{\alpha_i}$ is a shadow function, that is, $(\prod_{i=1}^{d+1}x_i^{\alpha_i})^{\Gamma}=0$. 
\end{lemma}

An important implication of this Lemma is that since each homogeneous polynomial in $\bigoplus_{i=0}^k\mathcal H_{2i+1}$ is a linear combination of monomials with at least one odd term, it is killed under $\pi_*$.  This implies that all ``odd'' pieces in $L^2(\mathbb S^d) =  \bigoplus_{i=0}^{\infty} \mathcal H_i$ do not contribute anything to  $L^2(\Delta^d)=L^2(\mathbb S^d/\Gamma)$. Therefore, whenever using spherical harmonics theory to understand function spaces of compositional domains, it suffices to consider only even $i$ for $\mathcal H_{i}$ in $L^2(\mathbb S^d)$. In summary, the function space on the compositional domain $\Delta^d=\mathbb S^d/\Gamma$ has the following eigenspace decomposition:
\begin{equation}\label{compfun}
 L^2(\Delta^d)=L^2(\mathbb S^d/\Gamma)=\dis\bigoplus_{i=0}^{\infty}\mathcal H_{2i}^{\Gamma},   
\end{equation}
\noindent where $\mathcal H_{2i}^{\Gamma}:=\{h\in \mathcal H_{2i},\ h=h^{\Gamma}\}$.

\subsection{Reproducing Kernels for Compositional Domain}\label{sec:rkhsc}

With the understanding of function spaces on compositional domains as invariant functions on spheres, we are ready to use spherical harmonic theory to construct reproducing kernel structures on compositional domains. 

\subsubsection{ $\Gamma$-invariant Functionals on $\mathcal H_i$}

%The identification $\Delta^d=\mathbb S^d/\Gamma$ helps us understand the function space $L^2(\Delta^d)=L^2(\mathbb S^d/\Gamma)$ as a subspace of $\Gamma$-invariant functions in $L^2(\mathbb S^d)$.  Reproducing kernels in $L^2(\mathbb S^d)$ are  well understood by the theory of spherical harmonics. Note that a compositional data point $z \in \Delta^d$ is identified as an orbit $\orb^{\Gamma}_z\subset \mathbb S^d$. 

The main goal of this section is to establish reproducing kernels for compositional data. Inside each Laplacian eigenspace $\mathcal H_i$ in $L^2(\mathbb S^d)$, recall that the $\Gamma$-invariant subspace $\mathcal H_i^{\Gamma}$ can be regarded as a function space on $\Delta^d=\mathbb S^d/\Gamma$, based on (\ref{compfun}).  To find a candidate of reproducing kernel inside $\mathcal H_i^{\Gamma}$, we first identify the representing function for the following linear functional $L_z^{\Gamma}$ on $\mathcal H_i$, which is defined as follows: For any function $Y\in \mathcal H_i$, 
\[
L_{z}^{\Gamma}(Y)=Y^{\Gamma}(z)=\dis\frac{1}{|\Gamma|}\sum_{\gamma\in \Gamma}Y(\gamma z),
\]
for a given $z\in \mathbb S^d$. One can easily see that $L_z^{\Gamma}$ and $L_z$ agree on the subspace $\mathcal H_i^{\Gamma}$ inside $\mathcal H_i$ and also that $L_z^{\Gamma}$ can be seen as a composed map $L_z^{\Gamma}=L_z\pi_*:\ \mathcal H_i\rightarrow \mathcal H_i^{\Gamma}\rightarrow\mathbb C$. Note that although $L_z^{\Gamma}$ is defined on $\mathcal H_i$, it can actually be seen as a ``Delta functional'' on $\mathbb S^d/\Gamma=\Delta^d$.

To find the representing function for $L_z^{\Gamma}$, we will use zonal spherical functions: Let $k_i(\cdot, \cdot)$ be the reproducing kernel in the eigenspace $\mathcal H_i$. Define the ``compositional'' kernel $k_i^{\Gamma}(\cdot, \cdot)$ for $\mathcal H_i$ as
\begin{equation}\label{fake}
   k_i^{\Gamma}(x, y)=\frac{1}{|\Gamma|}\sum_{\gamma\in \Gamma}k_i(\gamma x, y), \ \ \forall x, y\in \mathbb S^d,  
\end{equation}
\noindent from which it is straightforward to check that $k_i^\Gamma(z,\cdot)$ represents linear functionals of the form $L_{z}^{\Gamma}$,  simply by following the definitions. 

\begin{remark}
\normalfont The above definition of ``compositional kernels'' in (\ref{fake}) is not just a trick only to get rid of the ``redundant points'' on spheres. This definition is inspired by the notion of ``orbital integrals'' in analysis and geometry. In our case, the ``integral'' is a discrete version, because the ``compact subgroup'' in our situation is replaced by a finite discrete reflection group $\Gamma$. In fact, such kind of ``discrete orbital integral'' construction is not new in statistical learning theory, e.g., \cite{equimatr} also used the ``orbital integral'' type of construction to study equivariant matrix valued kernels.

\end{remark}

At first sight, a compositional kernel is not symmetric on the nose, because we are only ``averaging'' over the group orbit on the first variable of the function $k_i(x,y)$. However since $k_i(x,y)$ is both symmetric and orthogonally invariant by Propositional \ref{reprsph}, so quite counter-intuitively, compositional kernels are actually symmetric:

\begin{proposition}\label{sym}
 Compositional kernels are symmetric, namely $k_i^{\Gamma}(x, y)=  k_i^{\Gamma}(y, x)$. 
\end{proposition}

\begin{proof}
Recall that $k_i(x,y)=k_i(y,x)$ and that $k_i(G x,G y)=k_i(x,y)$ for any orthogonal matrix $G$. Notice that every reflection $\gamma\in \Gamma$ can be realized as an orthogonal matrix, then we have
$$
\begin{array}{rcl}
  k_i^{\Gamma}(x, y)&=&\dis\frac{1}{|\Gamma|}\sum_{\gamma\in \Gamma}k_i(\gamma x, y)\\
     &=&\dis\frac{1}{|\Gamma|}\sum_{\gamma\in \Gamma}k_i(y,\gamma x)=\frac{1}{|\Gamma|}\sum_{\gamma\in \Gamma}k_i(\gamma^{-1}y,\gamma^{-1}(\gamma x))\ \ \\
     &=&\dis\frac{1}{|\Gamma|}\sum_{\gamma\in \Gamma}k_i(\gamma^{-1}y,x)\\
     &=&\dis\frac{1}{|\Gamma|}\sum_{\gamma\in \Gamma}k_i(\gamma y,x)\\
     &=& k_i^{\Gamma}(y, x)
\end{array}
$$
\end{proof}

Recall that $\mathcal H_i^{\Gamma}$ is the $\Gamma$-invariant functions inside $\mathcal H_i$, and by (\ref{compfun}), $\mathcal H_i^{\Gamma}$ is the $i$-th subspace of a compositional function space $L^2(\Delta^d)$.  A na\"ive candidate for the reproducing kernel inside $\mathcal H_i^{\Gamma}$, denoted as $w_i(x,y)$, might be the spherical reproducing kernel $k_i(x,y)$, but $k_i(x,y)$ is not $\Gamma$-invariant. It turns out that the compositional kernels are actually reproducing with respect to all $\Gamma$-invariant functions in $\mathcal H_i$, while being $\Gamma$-invariant on both arguments.

\begin{theorem}\label{reprcomp}
Inside $\mathcal H_i$, the compositional kernel $k_i^{\Gamma}(x, y)$ is $\Gamma$-invariant on both arguments $x$ and $y$, and moreover $k_i^{\Gamma}(x, y)=w_i(x,y)$, i.e., the compositional kernel is the reproducing kernel for $\mathcal H^{\Gamma}_i$.
\end{theorem}
 
 \begin{proof}
Firstly by the definition, $k_i^{\Gamma}(x, y)$ is $\Gamma$-invariant on the first argument $x$; by the symmetry of  $k_i^{\Gamma}(x, y)$ in Proposition \ref{sym}, it is then also $\Gamma$-invariant on the second argument $y$, hence the compositional kernel $k_i^{\Gamma}(x, y)$ is a kernel inside $\mathcal H^{\Gamma}_i$.
 
Secondly, let us prove the reproducing property of $k_i^{\Gamma}(x, y)$. For any $\Gamma$-invariant function $f\in \mathcal H_i^{\Gamma}\subset \mathcal H_i$,
 $$
 \begin{array}{rcl}
     <f(t),k_i^{\Gamma}(x, t)> &=& <f(t),\dis\sum_{\gamma\in \Gamma}\frac{1}{|\Gamma|}k_i(\gamma x, t)>\\
     &=&\dis\frac{1}{|\Gamma|}\sum_{\gamma\in\Gamma}<f(t),k_i(\gamma x, t)>\\
     &=&\dis\frac{1}{|\Gamma|}\sum_{\gamma\in\Gamma}f(\gamma x)=\dis\frac{1}{|\Gamma|}\sum_{\gamma\in\Gamma}f(x)\ \ (f\ \text{is}\ \Gamma\text{-invariant})\\
      &=&f(x)
      
 \end{array}
 $$
 \end{proof}

\begin{remark}
\normalfont Theorem \ref{reprcomp} justifies that a compositional kernel is actually the reproducing kernel for functions inside $\mathcal H_i^\Gamma$. Although the compositional kernel $k_i^{\Gamma}(x, y)$ is symmetric as is proved in Proposition \ref{sym}, we will still use $w_i(x,y)$ to denote $k_i^{\Gamma}(x, y)$ because $w_i(x,y)$ is, notationally speaking, more visually symmetric than the notation for compositional kernels. 
\end{remark}

%Recall that functions on compositional domains are identified with $\Gamma$-invariant functions on $\mathbb S^d$ and points on compositional domains are $\Gamma$-orbits on the sphere, so reproducing kernels for $\Delta^d$ should reproduce $\Gamma$-invariant functions, as the following theorem shows. 

\subsubsection{Compositional RKHS and Spaces of Homogeneous Polynomials} 

Recall that based on Theorem \ref{accudirct}, the direct sum of an even (resp. odd)  number of eigenspaces can be expressed as the set of homogeneous polynomials of a fixed degree. Further recall that the direct sum decomposition $L^2(\mathbb S^d)=\bigoplus_{i=0}^{\infty}\mathcal H_i$ is an
orthogonal one, so is the direct sum $L^2(\Delta^d)=\bigoplus_{i=0}^{\infty}\mathcal H_i^\Gamma$.  By the orthgonality between eigenspaces, the reproducing kernels for the finite direct sum  $\bigoplus_{i=0}^m\mathcal H^{\Gamma}_i$ is naturally the summation $\sum_{i=0}^m w_i$. 
Note that by Lemma \ref{killodd}, it suffices to consider only even pieces of eigenspaces $\mathcal H_{2i}$. Finally, we give a formal definition of ``the degree $m$ reproducing kernel Hilbert space'' on $\Delta^d$, consisting degree $2m$ homogeneous polynomials:

\begin{definition}\label{lthrkhs}
Let $w_i$ be the reproducing kernel for $\Gamma$-invariant functions in the $i$th eigenspace $\mathcal H_i \subset L^2(\mathbb S^d)$. The degree $m$ compositional reproducing kernel Hilbert space is defined to be the finite direct sum $\bigoplus_{i=0}^{m}\mathcal H^{\Gamma}_{2i}$, and the reproducing kernel for the degree $m$ compositional reproducing kernel Hilbert space is

\begin{equation}\label{mcompker}
    \omega_m(\cdot, \cdot) = \sum_{i=0}^m w_{2i}(\cdot,\cdot). 
\end{equation}

\noindent Thus the degree $m$ RKHS for the compositional domain is the pair $\big(\bigoplus_{i=0}^{m}\mathcal H^{\Gamma}_{2i}, \omega_{m} \big)$. 
\end{definition}

Recall that the direct sum $\bigoplus_{i=0}^{m}\mathcal H^{\Gamma}_{2i}$ can identified as a subspace of $\bigoplus_{i=0}^{m}\mathcal H_{2i}$, which is isomorphic to the space of degree $2m$ homogeneous polynomials, so each function in $\bigoplus_{i=0}^{m}\mathcal H^{\Gamma}_{2i}$ can be written as a degree $2m$ homogeneous polynomial, including the reproducing kernel $\omega_m(x,\cdot)$, although it is not obvious from (\ref{mcompker}). Notice that for a point $(x_1,x_2,\dots, x_{d+1})\in \mathbb S^d$, the sum $\sum_{i=1}^{d+1}x_i^2=1$, so one can always use this sum to turn each element in $\bigoplus_{i=0}^{m}\mathcal H^{\Gamma}_{2i}$ to a homogeneous polynomial. For example, $x^2+1$ is not a homogeneous polynomial, but each point $(x,y,z)\in \mathbb S^2$ satisfies $x^2+y^2+z^2=1$, then we have $x^2+1=x^2+x^2+y^2+z^2=2x^2+y^2+z^2$, which is a homogeneous polynomial on the sphere $\mathbb S^2$.

In fact, we can say something more about $\bigoplus_{i=0}^{m}\mathcal H^{\Gamma}_{2i}$. Recall that Proposition \ref{killodd} ``killed'' the contributions from ``odd pieces'' $\mathcal H_{2k+1}$ under the  contraction map $\pi_*:\ L^2(\mathbb S^d)\rightarrow L^2(\Delta^d)$.  However, even inside $\bigoplus_{i=0}^{m}\mathcal H_{2i}$, only a subspace can be identified with a compositional function space, namely, those $\Gamma$-invariant homogeneous polynomials. The following proposition gives a characterization of which homogeneous polynomials inside $\bigoplus_{i=0}^{m}\mathcal H_{2i}$ come from the subspace $\bigoplus_{i=0}^{m}\mathcal H^{\Gamma}_{2i}$:

\begin{proposition}\label{mpolynomial}
Given any element $\theta\in \bigoplus_{i=0}^m\mathcal H^{\Gamma}_{2i}\subset \bigoplus_{i=0}^m\mathcal H_{2i}\subset L^2(\mathbb S^d/\Gamma)$, there exists a degree $m$ homogeneous polynomial $p_m$, such that 

\begin{equation}\label{mplynomial}
\theta(x_1, x_2,\dots, x_{d+1})=p_m(x_1^2, x_2^2,\cdots, x_{d+1}^{2}).
\end{equation}

\end{proposition}

\begin{proof}
Note that  $\theta$ is a degree $2m$ homogeneous $\Gamma$-invariant polynomial, then each monomial in $\theta$ has form $\prod_{i=1}^{d+1}x_i^{a_i}$ with $\sum_{i=1}^{d+1}a_i=2m$.

If $\theta$ contains one monomial $\prod_{i=1}^{d+1}x_i^{a_i}$ with nonzero coefficient such that $a_i$ is odd for some $1\leq i\leq d+1$. Note that $\theta$ is $\Gamma$-invariant, i.e., $\theta=\theta^{\Gamma}$, which implies $\prod_{i=1}^{d+1}x_i^{a_i}=(\prod_{i=1}^{d+1}x_i^{a_i})^{\Gamma}$, but the term $(\prod_{i=1}^{d+1}x_i^{a_i})^{\Gamma}$ is zero by Proposition \ref{killodd}. Thus $\theta$ is a linear combination of monomials of the form $\prod_{i=1}^{d+1}x_i^{a_i}=\prod_{i=1}^{d+1}(x_i^2)^{a_i/2}$ with each $a_i$ being even and $\sum_i a_i/2=m$, thus the proposition follows. 
\end{proof}

Recall that the degree $m$ compositional RKHS is $\big(\bigoplus_{i=0}^{m}\mathcal H^{\Gamma}_{2i}, \omega_{m} \big)$ in Definition \ref{lthrkhs}, and $\bigoplus_{i=0}^{m}\mathcal H_{2i}$ consists of degree $2m$ homogeneous polynomials while $\bigoplus_{i=0}^{m}\mathcal H^{\Gamma}_{2i}$ is just a subspace of it. Proposition \ref{mpolynomial} tells us that one can also have a concrete description of the subspace $\bigoplus_{i=0}^{m}\mathcal H^{\Gamma}_{2i}$ via those degree $m$ homogeneous polynomials on squared variables.

\section{Applications of Compositional Reproducing Kernels}\label{sec:app}

The availability of compositional reproducing kernels will open a door to many statistical/machine learning techniques for compositional data analysis. However, we will only present two application scenarios, as an initial demonstration of the influence of RKHS thoery on compositional data analysis. The first application is the representer theorem, which is motivated by newly developed kernel-based machine learning, especially by the rich theory of vector valued regression \citep{vecfun, vecman}. The second one is constructing exponential families on compositional domains. Parameters of compositional exponential models are compositional reproducing kernels.  To the best of authors' knowledge, these will be the first class of nontrivial examples of explicit distributions on compositional domains with \emph{non-vanishing} densities on the boundary.

\subsection{Compositional Representer Theorems}

Beyond the successful applications on traditional spline models, representer theorems are increasingly relevant due to the new kernel techniques in machine learning. We will consider minimal normal interpolations and least square regularzations in this paper. Regularizations are especially important in many situations, like structured prediction, multi-task learning, multi-label classification and related themes that attempt to exploit output structure.

A common theme in the above-mentioned contexts is non-parametric estimation of a vector-valued function $f:\ \mathcal X\rightarrow \mathcal Y$, between a structured input space $\mathcal X$ and a structured output space $\mathcal Y$. An important adopted framework in those analyses is the ``vector-valued reproducing kernel Hilbert spaces'' in \cite{vecfun}. Unsurprisingly, representer theorems not only are necessary, but also call for further generalizations in various contexts:

\begin{itemize}
    \item [(i)]In classical spline models, the most frequently used version of representer theorems are about scalar valued kernels, but besides the above-mentioned scenario  $f:\ \mathcal X\rightarrow \mathcal Y$ in manifold regularization context, in which vector valued representer theorems are needed, higher tensor valued kernels and their corresponding representer theorems are also desirable. In \cite{equimatr}, matrix valued kernels and their representer theorems are studied, with applications in image processing. 
    
    \item[(ii)]Another related application lies in the popular kernel mean embedding theories, in particular, conditional mean embedding. Conditional mean embedding theory essentially gives an operator from an RKHS to another \citep{condmean}. In order to learn such operators, vector-valued regressions plus corresponding representer theorems are used.  
\end{itemize}

In vector-valued regression framework, an important assumption discussed in representer theorems are \emph{linear independence} conditions \citep{vecfun}. As our construction of compositional RKHS is based on finite dimensional spaces of polynomials, the linear independence conditions are not freely satisfied on the nose, so we will address this problem in this paper. Instead of dealing with vector-valued kernels, we will only focus on the special case of scalar valued (reproducing) kernels, but the issue can be clearly seen in this special case.

\subsubsection{Linear Independence of Compositional Reproducing Kernels}\label{twist}

The compositional RKHS that was constructed in Section \ref{sec:rkhs} takes the form $\big(\bigoplus_{i=0}^{m}\mathcal H_{2i}^\Gamma, \omega_{m} \big)$ indexed by $m$. Based on the finite dimensional nature of compositional RKHS,  it is not even clear whether different points yield to different functions $\omega_{m}(x_i, \cdot)$ inside $\bigoplus_{i=0}^{m}\mathcal H_{2i}^\Gamma$. we will give a positive answer when $m$ is high enough.

Given a set of distinct compositional data points $\{x_i\}_{i=1}^n\subset \Delta^d$, we will show that the corresponding set of reproducing functions $\{\omega_{m}(x_i, \cdot)\}_{i=1}^n$ form a linearly independent set inside $\bigoplus_{i=0}^{m}\mathcal H_{2i}^\Gamma$ if $m$ is high enough.

\begin{theorem}\label{abundencelm} Let $\{x_i\}_{i=1}^n$ be distinct data points on a compositional domain $\Delta^d$. Then there exists a positive integer $M$, such that for any $m>M$, the set of functions $\omega_{m}(x_i, \cdot)\}_{i=1}^n$ is a linearly independent set in $\bigoplus_{i=0}^{m}\mathcal H_{2i}^\Gamma$.

\end{theorem}

\begin{proof} 

The quotient map $c_\Delta: \mathbb S^d\rightarrow \Delta^d$ can factor through a projective space, i.e., $c_\Delta:\ \mathbb S^d\rightarrow\mathbb P^d\rightarrow \Delta^d$.  The main idea is to prove a stronger statement, in which we showed that distinct data points in $\mathbb P^d$ will give linear independence of \emph{projective kernels} for large enough $m$,  where projective kernels are reproducing kernels in $\mathbb P^d$ whose definition was given in \ref{abundprf}. Then we construct two  vector subspace $V_1^{m}$ and $V_2^{m}$ and a linear map $g_m$ from $V_1^{m}$ to $V_2^{m}$. The key trick is that the matrix representing the linear map $g_m$ becomes diagonally dominant when $m$ is large enough, which forces the spanning sets of both $V_1^{m}$ and $V_2^{m}$ to be linear independent. More details of the proof are given in Section \ref{abundprf}.

\end{proof}

In the proof of Theorem \ref{abundencelm}, we make use of the homogeneous polynomials $(y_i\cdot {t})^{2m}$, which is \emph{not} living inside a single piece $\mathcal H_{2i}$, thus we had to use the direct sum space $\bigoplus_{i=0}^{m}\mathcal H_{2i}$ for our construction of RKHS.  Without using projective kernels, one might wonder if the same argument works, however, the issue is that the matrix might have $\pm 1$ at off-diagonal entries, which will fail to be diagonally dominant when $m$ grows large enough.  We break down to linear independence of projective kernels for distinct points, because reproducing kernels for distinct compositional data points are linear combinations of distinct projective kernels, then in this way, the off diagonal terms will be the power of inner product of two vectors that will not be antipodal or identical, thus off diagonal terms' $m$-th power will go to zero with increasing $m$. 

Another consequence of Theorem \ref{abundencelm} is that $\omega_{m}(x_i,\cdot)\neq \omega_{m}(x_j, \cdot)$ whenever $i\neq j$ when $m$ is large enough. Not only large enough $m$ will separate points on the reproducing kernel level, but also gives each data point their ``own dimension.''

\subsubsection{Minimal Norm Interpolation and Least Squares Regularization}\label{represent}

Once the linear independence is established in Theorem \ref{abundencelm}, it is an easy corollary to establish the representer theorems for minimal norm interpolations and least square regularizations. Nothing is new from the point of view of general RKHS theory, but we will include these theorems and proofs on account of completeness. Again, we will focus on the scalar-valued (reproducing) kernels and functions, instead of the vector-valued kernels and regressions. However, Theorem \ref{abundencelm} sheds important lights on linearly independence issues, and interested readers can generalize these compositional representer theorems to vector-valued cases by following \cite{vecfun}. 

The first representer theorem we provide is a solution to minimal norm interpolation problem:  for a fixed set of distinct points $\{x_i\}_{i=1}^n$ in $\Delta^d$ and a set of numbers $y=\{y_i\in \mathbb R\}_{i=1}^n$, let $I_y^{m}$ be the set of functions that interpolates the data
\[
I_y^m = \{f\in \bigoplus_{i=0}^{m}\mathcal H^{\Gamma}_{2i}:\ f(x_i)=y_i\},
\]
and out goal is to find $f_0$ with minimum $\ell_2$ norm, i.e., 
\[
\norm{f_0}=\inf\{\norm{f}, f\in I_y^{m}\}.\]

\begin{theorem}\label{minorm}
Choose $m$ large enough so that the reproducing kernels $\{\omega_m(x_i,t)\}_{i=1}^n$ are linearly independent, then the unique solution of the minimal norm interpolation problem $\min\{\norm{f},f\in  \bigoplus_{i=0}^{m}\mathcal H^{\Gamma}_{2i}:\ f(x_i)=y_i\}$ is given by the linear combination of the kernels:
$$
f_0(t)=\dis\sum_{i=1}^nc_i \; \omega_{m}(x_i,t)
$$
where $\{c_i\}_{i=1}^n$ is the unique solution of the following system of linear equations:
$$
\dis\sum_{j=1}^n\omega_{m}(x_i,x_j)c_j=y_i, \ \ 1\leq i\leq n.
$$
\end{theorem}

\begin{proof}
For any other $f$ in $I_y^{m}$, define $g=f-f_0$. By considering the decomposition: $\norm{f}^2=\norm{g+f_0}^2=\norm{g}^2+2<f_0,g>+\norm{f_0}^2$,  one can argue that the cross term $<f_0,g>=0$. The detail can be found in Section \ref{representerproofs}. We want to point out that the linear independence of reproducing kernels guarantees the uniqueness and existence of $f_0$.
\end{proof}

The second representer theorem is for a more realistic scenario with $\ell_2$ regularization, which has the following objective: 
\begin{equation}\label{l2obj}
 \sum_{i=1}^n  \left ( f(x_i)-y_i \right )^2+\mu\norm{f}^2.
\end{equation}
The goal is to find the $\Gamma$-invariant function $f_{\mu}\in  \bigoplus_{i=0}^{m}\mathcal H^{\Gamma}_{2i}$ that minimizes (\ref{l2obj}).  The solution to this problem is provided by the following representer theorem:

\begin{theorem}\label{regularization}
For a set of distinct compositional data points $\{x_i\}_{i=1}^n$, choose $m$ large enough such that the reproducing kernels 
$\{\omega_{m}(x_i, t)\}_{i=1}^n$ are linearly independent. Then the solution to (\ref{l2obj}) is given by 
\[
f_{\mu}(t) =\sum_{i=1}^n c_i \; \omega_{m}(x_i,t),
\]
where $\{c_i\}_{i=1}^n$ is the solution of the following system of linear equations:
\[
\mu c_i+\sum_{j=1}^n  \omega_{m}(x_i,x_j)c_j=y_i,\ \ 1\leq i\leq n.
\]
\end{theorem}

\begin{proof} The detail of this proof can be found in Section \ref{representerproofs}, but we want to point out how the linear independence condition plays a role in here. In the middle of the proof we need to show that $\mu f_{\mu}(t)=\sum_{i=1}^n\big[(y_i-f_{\mu}(x_i)) \omega_{m}(x_i,t)\big]$, where  $f_{\mu}(t) =\sum_{i=1}^n \omega_{m}(x_i,t)c_i$. We use the linear independence in Theorem \ref{abundencelm} to establish the equivalence between this linear equation system of $\{c_i\}_{i=1}^n$ and the one given in the theorem.
\end{proof}

\subsection{Compositional Exponential Family}

With the construction of RKHS in hand, one can produce exponential families using the technique  developed in \cite{cs06}.  
Recall that for a function space $\mathcal H$ with the inner product $<\cdot,\cdot>$ on a general topological space $\mathcal X$, whose reproducing kernel is given by $k(x, \cdot)$, the exponential family density $p(x, \theta)$ with the parameter $\theta\in \mathcal H$ is given by:
\[
p(x, \theta)=\exp\{<\theta(\cdot),k(x,\cdot)>-g(\theta) \},\ 
\]
where $g(\theta)=\log \dis\int_{\mathcal{X}}\exp\big(<\theta(\cdot),k(x,\cdot) >\big) dx$. 

For compositional data we define the density of the $m$th degree exponential family as

\begin{equation}\label{expfamily}
p_{m}(x, \theta)=\exp\left\{<\theta(\cdot),\omega_{m}(x,\cdot)>-g(\theta) \right \},\quad x \in \mathbb S^d/\Gamma,
\end{equation}
where $\theta\in \bigoplus_{i=0}^{m}\mathcal H^{\Gamma}_{2i}$ and $g(\theta)=\log \int_{\mathbb S^d/\Gamma}\exp(<\theta(\cdot),\omega_{m}(x,\cdot) >) dx$. Note that  this density can be made more explicit by using homogeneous polynomials. Recall that any function in $\bigoplus_{i=0}^m\mathcal H_{2i}^{\Gamma}$ can be written as a degree $m$ homogeneous polynomial with \emph{squared} variables by Lemma \ref{killodd}. Thus the density in (\ref{expfamily}) can be simplified to the following form: for $x=(x_1,\dots, x_{d+1})\in \mathbb S^{d}_{\geq 0}$,
 \begin{equation}\label{mpoly}
  p_{m}(x,\theta)=\exp\{s_{m}(x_1^2, x_2^2, \dots, x_{d+1}^2; \theta)-g(\theta)\},
 \end{equation} 
where $s_m$ is a polynomial on squared variables $x_i^2$'s with $\theta$ as coefficients. Note that $s_m$ is invariant under ``sign-flippings'', and the normalizing constant can be computed via the integration over the entire sphere as follows:
 
  $$
 g(\theta)=\int_{\mathbb S^d/\Gamma}\exp(s_m)dx=\frac{1}{|\Gamma|}\int_{\mathbb S^d}\exp(s_m)dx.
 $$

\begin{figure}[h]
\begin{subfigure}[b]{0.32\textwidth}
\includegraphics[width = \textwidth]{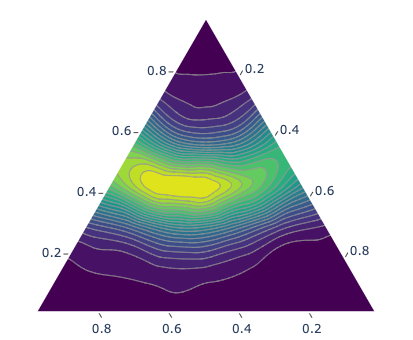}
\caption{$p_4(x,\theta_1)$}
\end{subfigure}
\begin{subfigure}[b]{0.32\textwidth}
\includegraphics[width = \textwidth]{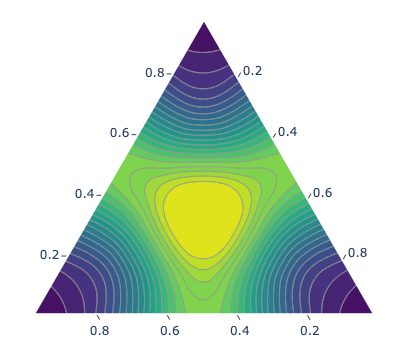}
\caption{$p_4(x,\theta_2)$}
\end{subfigure}
\begin{subfigure}[b]{0.32\textwidth}
\includegraphics[width = \textwidth]{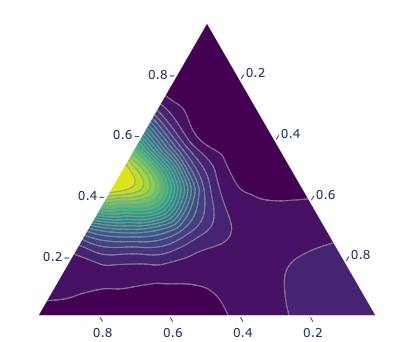}
\caption{$p_4(x,\theta_3)$}
\end{subfigure}
\caption{Three example densities from compositional exponential family. See text for specification of the parameters $\theta_1, \theta_2$, and $\theta_3$. }\label{fig:expfamily}
\end{figure}

Figure \ref{fig:expfamily} displays three examples of compositional exponential distribution. The three densities respectively have the following $\theta$:
\begin{eqnarray*}
\theta_1 &=&  -2 x_1^4 -2 x_2^4  -3 x_3^4 + 9 x_1^2 x_2^2 + 9 x_1^2 x_3^2 -2 x_2^2 x_3^2,\\
\theta_2 &=&  - x_1^4 - x_2^4  - x_3^4 - x_1^2 x_2^2 - x_1^2 x_3^2 - x_2^2 x_3^2,\\
\theta_3 &=&  - 3 x_1^4 - 2 x_2^4  - x_3^4 +9 x_1^2 x_2^2 - 5 x_1^2 x_3^2 - 5 x_2^2 x_3^2.
\end{eqnarray*}
The various shapes of the densities in the Figure implies that the compositional exponential family can be used to model data with a wide range of locations and correlation structures. 

Further investigation is needed on the estimation of the parameters of the compositional exponential model (\ref{mpoly}), which is suggested as a future direction of research. A natural starting point is maximum likelihood estimation and a regression-based method such as the one discussed by \cite{expdirect}.

\section{Discussion}\label{sec:conclude}

A main contribution of this work is that we use projective and spherical geometries to reinterpret compositional domains, which allows us to construct reproducing kernels for compositional data points by using spherical harmonics under group actions. With the rapid development of kernel techniques (especially kernel mean embedding philosophy) in machine learning theory, this work will make it possible to introduce reproducing kernel theories to compositional data analysis. 

Let us for example consider the mean estimation problem for compositional data. Under the kernel mean embedding framework that is surveyed by \citet{kermean}, one can focus on kernel mean $\mathbb E[k(X,\cdot)]$ in the function space, rather a physical mean that exist in the compositional domain. The latter is known to be difficult to even define  properly \citep{sphkentreg, sw11}. On the other hand, the kernel mean is endowed with flexibility and linear structure of the function space. Although inspired by the kernel mean embedding techniques, we did not address the computation of the kernel mean $\mathbb E[k(X,\cdot)]$ and cross-variance operator as a replacement of traditional means and variance-covariances in traditional multivariate analysis. The authors will, in the forthcoming work,  develop further techniques to come back to this issue of kernel means and cross-variance operators for compositional exponential models, via applying deeper functional analysis techniques. 

Although we only construct reproducing kernels for compositional data, it does not mean that ``higher tensors'' is abandoned in our consideration. In fact, higher-tensor valued reproducing kernels are also included in kernel techniques with applications in manifold regularizations \citep{vecman} and shape analysis \citep{matrixvaluedker}. These approaches on higher-tensor valued reproducing kernels indicate further possibilities of regression frameworks between exotic spaces $f:\ \mathcal X\rightarrow \mathcal Y$ with both the source $\mathcal X$ and $\mathcal Y$ being non-linear in nature, which extends the intuition of multivariate analysis further to nonlinear contexts, and compositional domains (traditionally modeled by an ``Aitchison simplex'') are an interesting class of examples which can be treated non-linearly.

\appendix

\section{Supplementary  Proofs}

\subsection{ Proof of Central Limit Theorems on Integral Squared Errors (ISE) in Section \ref{compdensec}}\label{cltprf}

\begin{assumption}\label{kband}
For all kernel density estimators and bandwidth parameters in this paper, we assume the following:

\begin{itemize}
\item[{\bf{H1}}] The kernel function $K:[0,\infty)\rightarrow [0,\infty)$ is continuous such that both $\lambda_d(K)$ and $\lambda_d(K^2)$ are bounded for $d\geq 1$, where $\lambda_d(K)=2^{d/2-1}\mathrm{vol}(S^d)\dis\int_{0}^{\infty}K(r)r^{d/2-1}dr$.

\item[\bf{H2}]If a function $f$ on $\mathbb S^d\subset \mathbb R^{d+1}$ is extended to the entire $\mathbb R^{d+1}/\{0\}$ via $f(x)=f(x/\norm{x})$, then the extended function $f$ needs to have its first three derivatives bounded.

\item[\bf{H3}] Assume the bandwidth parameter $h_n\rightarrow 0$ as $nh_n^d\rightarrow\infty$.
\end{itemize}

\end{assumption}

Let $f$ be the extended function from $\mathbb S^d$ to $\mathbb R^{d+1}/\{0\}$ via $f(x/\norm{x})$, and let 
$$
\phi (f,x)= -x^T \nabla f(x)+d^{-1}(\nabla^2 f(x)-x^T (\mathcal H_x f) x) = d^{-1}\mathrm{tr} [\mathcal H_xf(x)],
$$
where $\mathcal H_x f$ is the Hessian matrix of $f$ at the point $x$.  

The term $b_d(K)$ in the statement of Theorem \ref{ourclt} is defined to be:
\[
b_d(K)=\dis\frac{\dis\int_0^{\infty}K(r)r^{d/2}dr}{\dis\int_0^{\infty}K(r)r^{d/2-1}dr}
\]

The term $\phi(h_n)$ in the statement of Theorem \ref{ourclt} is defined to be:
\[
\phi(h_n)=\dis\frac{4b_d(K)^2}{d^2}\sigma_x^2h_n^4
\]

Proof of Theorem \ref{ourclt}:
\begin{proof}
The strategy in \cite{chineseclt} in the directional set-up follows that in \cite{hallclt}, whose key idea is to give asymptotic bounds for degenerate U-statistics, so that one can use Martingale theory to derive the central limit theorem.  The step where the finite support condition was used in \cite{chineseclt}, is when they were trying to prove the asymptotic bound:$E(G_n^2(X_1,X_2))=O(h^{7d})$, where $G_n(x,y)=E[H_n(X,xH_n(X,y))]$ with $H_n=\dis\int_{\mathbb S^d}K_n(z,x)K_n(z,y)dz$ and the centered kernel $K_n(x,y)=K[(1-x'y)/h^2]-E\{[K(1-x'X)/h^2]\}$. During that prove, they were trying to show that the following term:

\[
\begin{array}{rcl}
T_1&=&\dis\int_{\mathbb S^d}f(x)dx\int_{\mathbb S^d}f(y)dy\times\\
   &&\left\{
\dis\int_{\mathbb S^d}f(z)dz\dis\int_{\mathbb S^d}K[(1-u'x)/h^2]K[(1-u'z)/h^2]du \cdot
\int_{\mathbb S^d}K[(1-u'y)/h^2]K[(1-u'z)/h^2]du
\right\}^2,
\end{array}
\]

satisfies $T_1=O(h^{7d})$. During this step, in order to give an upper bound for $T_1$, the finite support condition was substantially used.  

The idea to avoid this assumption was based on the observation in \cite{portclt} where they only concern the case of directional-linear CLT, whose result can not be directly used to the only directional case.  Based on the method provided in Lemma 10 in \cite{portclt}, one can easily deduce the following asymptotic equivalence:
$$
\dis\int_{\mathbb S^d}K^j(\frac{1-x^Ty}{h^2})\phi^i(y)dy\sim h^d\lambda_d(K^j)\phi^i(x),
$$
where $\lambda_d(K^j)=2^{d/2-1}\mathrm{vol}(\mathbb S^{d-1})\dis\int_{0}^{\infty}K^j(r)r^{d/2-1}dr$. As a special case we have:
$$
\dis\int_{\mathbb S^d}K^2(\dis\frac{1-x^Ty}{h^2})dy\sim h^d\lambda_d(K^2)C,\ \text{with}\ C\ \text{being a positive constant}.
$$

Now we will proceed the proof without the finite support condition:
$$
\begin{array}{rcl}
T_1&=&\dis\int_{\mathbb S^d}f(x)dx\int_{\mathbb S^d}f(y)dy\\
   &&\times\left\{
\dis\int_{\mathbb S^d}f(z)dz\dis\int_{\mathbb S^d}K[(1-u'x)/h^2]K[(1-u'z)/h^2]du \cdot
\int_{\mathbb S^d}K[(1-u'y)/h^2]K[(1-u'z)/h^2]du
\right\}^2\\
   &\sim& \dis\int_{\mathbb S^d}f(x)dx\int_{\mathbb S^d}f(y)dy\left\{\int_{\mathbb S^d}f(z)\big[\lambda_d(K)h^dK(\frac{1-x^Tz}{h^2})\big]\times \big[\lambda_d(K)h^dK(\dis\frac{1-y^Tz}{h^2})\big]dz \right\}^2\\
   &\sim& \lambda_d(K)^4h^{4d}\dis\int_{\mathbb S^d}f(x)dx\int_{\mathbb S^d}f(y)\big[ \lambda_d(K)h^d K(\dis\frac{1-x^Ty}{h^2})f(y)\big]^2dy\\
   &=&\lambda_d(K)^6h^{6d}\dis\int_{\mathbb S^d} \left\{\int_{\mathbb S^d} K^2(\dis\frac{1-x^Ty}{h^2})f^3(y)dy \right\}f(x)dx\\
   &\sim&\lambda_d(K)^6h^{6d}\dis\int_{\mathbb S^d} \lambda_d(K^2)h^dC\cdot f^3(x)f(x)dx\\
   &=&C\lambda_d(K)^6\lambda_d(K^2)h^{7d}\dis\int_{\mathbb S^d}f^(x)dx=O(h^{7d}).
\end{array}
$$
Thus we have proved $T_1=O(h^{7d})$ without finite support assumption, then the rest of the proof will follow through as in \cite{chineseclt}.
\end{proof}

Observe the identity:
\begin{equation}\label{csden}
\dis\int_{\mathbb S^d_{\geq 0}}(\hat{p}_n-p)^2dx=\dis|\Gamma|\int_{\mathbb S^d}(\hat{f}_n-\tilde{p})^2dy,
\end{equation}

then the CLT of compositional ISE follows from the identity (\ref{csden}) and our proof of CLT on spherical ISE without finite support conditions on kernels.

\subsection{Proofs of Shadow Monomials in Section \ref{sec:rkhs} }\label{reproproof}

Proof of Proposition \ref{killodd}:

\begin{proof}
A direct computation yields:
$$
\begin{array}{rcl}
(\prod_{i=1}^{d+1}x_i^{\alpha_i})^{\Gamma}&=&\dis\frac{1}{|\Gamma|}\sum_{s_i\in \{\pm 1\}}\prod_{i=1}^{d+1}(s_ix_i)^{\alpha_i}\\
   &=&\dis\frac{1}{|\Gamma|}\sum_{s_i\in \{\pm 1\}}\prod_{i\neq k}(s_ix_i)^{\alpha_i} x_k^{\alpha_k}+\sum_{s_i\in \{\pm 1\}}\prod_{i\neq k}(s_ix_i)^{\alpha_i} (-x_k)^{\alpha_k}\\
   &=&x_k^{\alpha_k}\dis\frac{1}{|\Gamma|}\sum_{s_i\in \{\pm 1\}}\prod_{i\neq k}(s_ix_i)^{\alpha_i}-x_k^{\alpha_k}\dis\frac{1}{|\Gamma|}\sum_{s_i\in \{\pm 1\}}\prod_{i\neq k}(s_ix_i)^{\alpha_i} \\
   &=&0.
\end{array}
$$

\end{proof}

\subsection{Proof of Linear Independence of Reproducing Kernels in Theorem \ref{abundencelm}}\label{abundprf}

We sketch a slight of more detailed (not complete) proof:

\begin{proof}
 This is the most technical lemma in this article. We will sketch the philosophy of the proof in here, which can be intuitively understood topologically.

Recall that we can produce a projective space $\mathbb P^d$ by identifying every pair of antipodal points of a sphere $\mathbb S^d$ (identify $x$ with $-x$), in other words $\mathbb P^d=\mathbb S^d/\mathbb Z_2$ where $\mathbb Z_2=\{0,1\}$ is a cyclic group of order $2$.  Then we can define a projective kernel in $\mathcal H_i\subset L^2(\mathbb S^d)$ to be $k^p_i(x, \cdot)=[k_i(x,\cdot)+k_i(-x,\cdot)]/2$.  We can also denote the projective kernel inside $\bigoplus_{i=0}^{m}\mathcal H_{2i}$ by $\underline{k}_m^p(x, \cdot)=\sum_{i=0}^{m}k^p_{2i}(x,\cdot)$. 

Now we spread out the data set $\{x_i\}_{i=1}^n$ by ``spread-out'' construction in Section \ref{sec:spread}, and denote the spread-out data set as $\{\Gamma \cdot x_i\}_{i=1}^n=\{{c^{-1}_{\Delta}(x_i)}\}_{i=1}^{n}$ (a data set, not a set because of repetitions).  A compositional reproducing kernel kernel is a summation of spherical reproducing kernels of on ${c^{-1}_{\Delta}(x_i)}$, divided by the number of elements in ${c^{-1}_{\Delta}(x_i)}$. This data set ${c^{-1}_{\Delta}(x_i)}$ has antipodal symmetry, then a compositional kernel is a linear combination of projective kernels. Notice that \emph{different} fake kernels are linear combinations of \emph{different} projective kernels. It suffices to show the linear independence of projective kernels for distinct data points and large enough $m$, which implies the linear independence of fake kernels $\{\underline{k}^{\Gamma}_{m}(x_i, \cdot)\}_{i=1}^n$.

Now we are focusing on the linear independence of projective kernels. A projective kernel can be seen as a reproducing kernel for a point in $\mathbb P^d$. For a set of distinct points $\{y_i\}_{i=1}^l\subset \mathbb P^{d}$, we will show that the corresponding set of projective kernels $\{\underline{k}^{p}_{m}(y_i, \cdot)\}_{i=1}^l\subset \bigoplus_{i=0}^{m}\mathcal H_{2i}$ is linearly independent for an integer $l$ and a large enough $m$.

Consider two vector subspace $V_1^{m}=\spn \big[ \{(y_i\cdot {t})^{2m}\}_{i=1}^l\big]$ and $V_2^{m}=\spn\big[\{\underline{k}^{p}_{m}(y_i, t)\}_{i=1}^l\big]$, both of which are  inside $\bigoplus_{i=0}^{m}\mathcal H_{2i}\subset L^2(\mathbb S^d)$.  Then we can define a linear map $h_{m}:\ V_1^{m}\rightarrow V_2^{m}$ by setting $h_{m}((y_i\cdot {t})^{2m})=\sum_{j=1}^l<(y_i\cdot {t})^{2m},\underline{k}^{p}_{m}(y_j, t)>\underline{k}^{p}_{m}(y_j, t)$.  This linear map $h_{m}$ is represented by an $l\times l$ symmetric matrix whose diagonal elements are $1$'s, and off diagonal elements are $[(y_i\cdot y_j)]^{2m}$. Notice that $y_i\neq y_j$ in $\mathbb P^d$, which means that they are not antipodal to each other in $\mathbb S^d$, thus $|y_i\cdot y_j|<1$.  When $m$ is large enough, all off-diagonal elements will go to zero while diagonal elements always stay constant, then the matrix representing $h_{m}$ will become a \emph{diagonally dominant} matrix, which is full rank.  When the linear map $h_{m}$ has full rank, both spanning sets $ \{(y_i\cdot {t})^{2m}\}_{i=1}^l$ and $\{\underline{k}^{p}_{m}(y_i, t)\}_{i=1}^l$ have to be a basis for $V_1^{m}$ and $V_2^{m}$ correspondingly, then the set of projective kernels $\{\underline{k}^{p}_{m}(y_i, t)\}_{i=1}^m$ have to be linearly independent when $m$ is large enough.

\end{proof}

\subsection{Proof of Representer Theorems in Section \ref{represent}}\label{representerproofs}

Proof of Theorem \ref{minorm} on minimal norm interpolation:

\begin{proof}
Note that the set $I_y^{m}=\{f\in  \bigoplus_{i=0}^{m}\mathcal H_{2i}:\  f(x_i)=y_i\}$ is non-empty, because the $f_0$ defined by the linear system of equation is naturally in $I_y^{m}$. Let $f$ be any other element in $I_y^{m}$, define $g=f-f_0$, then we have:
$$
\norm{f}^2=\norm{g+f_0}^2=\norm{g}^2+2<f_0,g>+\norm{f_0}^2.
$$ 
Notice that $g\in  \bigoplus_{i=0}^{m}\mathcal H_{2i}$ and that $g(x_i)=0$ for $1\leq i\leq n$, we have:
$$
\begin{array}{rcl}
<f_0,g>&=&<\sum_{i=1}^n\omega_{m}(x_i,\cdot)c_i,g(\cdot)>\\
            &=&\dis\sum_{i=1}^nc_i<\omega_{m}(x_i,\cdot),g(\cdot)>.\\
            &=&\dis\sum_{i=1}^nc_ig(x_i)=0.
\end{array}
$$
Thus $\norm{f}^2=\norm{g+f_0}^2=\norm{g}^2+\norm{f_0}^2$, which implies that $f_0$ is the solution to the minimal norm interpolation problem.
\end{proof}

Proof of Theorem \ref{regularization}, on regularization problems:

\begin{proof}
First define the loss functional $E(f)=\sum_{i=1}^n|f(x_i)-y_i|^2+\mu\norm{f}^2$. For any $\Gamma$-invariant function $f=f^{\Gamma}\in \bigoplus_{i=0}^{m}\mathcal H_{2i}$, let $g=f-f_{\mu}$, then a simple computation yields:
$$
\dis E(f)=E(f_{\mu})+\sum_{i=1}^n |g(x_i)|^2 -2\sum_{i=1}^n(y_i-f_{\mu}(x_i))g(x_i)+2\mu<f_{\mu},g>+\mu\norm{g}^2.
$$
I want to show $\sum_{i=1}^n(y_i-f_{\mu}(x_i))g(x_i)=\mu<f_{\mu},g>$, and an equivalent way of writing this equality is:
$$
\dis\sum_{i=1}^n<(y_i-f_{\mu}(x_i))\omega_{m}(x_i,t), g(t)>=\mu<f_{\mu},g>.
$$
Now I claim that $\mu f_{\mu}(t)=\sum_{i=1}^n\big[(y_i-f_{\mu}(x_i))\cdot \omega_{m}(x_i,t)\big]$, which implies the above equality.  To prove this claim, plug this linear combination $f_{\mu}=\sum_{i=1}^nc_i\cdot \omega_{m}(x_i,t)$ into the claim, then we get a system of linear equations in $\{c_i\}_{i=1}^n$, thus the proof of the claim breaks down to checking the system of linear equations in $\{c_i\}_{i=1}^n$, produced by the claim.

Note that $\{\omega_{m}(x_i,t)\}_{i=1}^n$ is a linearly independent set, so one can check that the system of linear equations in $\{c_i\}_{i=1}^n$ produced by the claim is true, if and only if $\{c_i\}_{i=1}^n$ satisfy
 $\mu c_k+\sum_{i=1}^n c_i\cdot \omega_{m}(x_i,x_k)=y_k$ for every $k$ with $1\leq k\leq n$, which is given by the condition of this theorem.  The equivalence of these two systems of linear equations is given by the linear independence of the set $\{\omega_{m}(x_i, t)\}_{i=1}^n$.  Therefore we conclude that the claim  $\mu f_{\mu}(t)=\sum_{i=1}^n\big[(y_i-f_{\mu}(x_i))\cdot  \omega_{m}(x_i,t)\big]$ is true.
 
  To finish the proof of this theorem, notice that
 $$
 \begin{array}{rcl}
 \dis E(f)&=&E(f_{\mu})+\sum_{i=1}^n |g(x_i)|^2 -2\sum_{i=1}^n(y_i-f_{\mu}(x_i))g(x_i)+2\mu<f_{\mu},g>+\mu\norm{g}^2\\
             &=&\dis E(f_{\mu})+\sum_{i=1}^n |g(x_i)|^2+\mu\norm{g}^2+2\big[\mu<f_{\mu},g>-\sum_{i=1}^n(y_i-f_{\mu}(x_i))g(x_i)\big]\\
             &=&\dis E(f_{\mu})+\sum_{i=1}^n |g(x_i)|^2+\mu\norm{g}^2+2\big[\mu<f_{\mu},g>-\sum_{i=1}^n<(y_i-f_{\mu}(x_i))\cdot \omega_{m}(x_i,t), g(t)>\big]\\
             &=&\dis E(f_{\mu})+\sum_{i=1}^n |g(x_i)|^2+\mu\norm{g}^2+2\big[<\underbrace{\big(\mu f_{\mu}(t)-\sum_{i=1}^n\big[(y_i-f_{\mu}(x_i))\cdot \omega_{m}(x_i,t)\big]\big)}_{=0}, g(t)>\big]\\
             &=&\dis E(f_{\mu})+\sum_{i=1}^n |g(x_i)|^2+\mu\norm{g}^2.
 \end{array}
 $$
 The term $\sum_{i=1}^n |g(x_i)|^2+\mu\norm{g}^2$ in the above equality is always non-negative, thus $E(f_{\mu})\leq   E(f)$, then the theorem follows.
\end{proof}

\bibliographystyle{asa}
\bibliography{geometrycomp.bib}

\end{document}